\documentclass[sigconf]{acmart}

\usepackage[english]{babel}
\usepackage{blindtext}
\usepackage{caption}
\usepackage{subcaption}
\usepackage{color}
\usepackage{tabularx}
\usepackage{multirow}
\usepackage{multicol}
\usepackage{graphicx}
\usepackage{tabularray}
\usepackage{float}
\usepackage{xcolor} 
\usepackage{array} 
\usepackage{makecell} 
\usepackage{booktabs} 

\newcommand{\name}{MMBind\xspace}

\def\thickhline{\noalign{\hrule height.8pt}}

\begin{document}

\acmYear{2025}\copyrightyear{2025}
\acmConference[SenSys '25]{The 23rd ACM Conference on Embedded Networked Sensor Systems}{May 6--9, 2025}{Irvine, CA, USA}
\acmBooktitle{The 23rd ACM Conference on Embedded Networked Sensor Systems (SenSys '25), May 6--9, 2025, Irvine, CA, USA}
\acmDOI{10.1145/3715014.3722053}
\acmISBN{979-8-4007-1479-5/25/05}

\title{\name: Unleashing the Potential of Distributed and Heterogeneous Data for Multimodal Learning in IoT}

\author{Xiaomin Ouyang$^{1, * \dagger}$, Jason Wu$^{2, \dagger}$, Tomoyoshi Kimura$^{3}$, Yihan Lin$^{2}$, Gunjan Verma $^{4}$, Tarek Abdelzaher$^{3}$, Mani Srivastava$^{2, \ddagger}$
}

\affiliation{  
\institution{$^{1}$Hong Kong University of Science and Technology, $^{2}$University of California, Los Angeles, $^{3}$University of Illinois Urbana-Champaign, $^{4}$U.S. Army DEVCOM Army Research Laboratory}
\country{}
}

\thanks{*The work was done when the author was at UCLA. $\dagger$Both authors contributed equally to this research. $\ddagger$ The author holds concurrent appointments as Amazon Scholar and Professor at UCLA, but work in this paper is not associated with Amazon.}









\renewcommand{\shortauthors}{X. Ouyang, J. Wu, T. Kimura, Y. Lin, G. Verma, T. Abdelzaher, M. Srivastava}

\begin{abstract}
Multimodal sensing systems are increasingly prevalent in various real-world applications. Most existing multimodal learning approaches heavily rely on training with \emph{a large amount of synchronized, complete multimodal data}. However, such a setting is impractical in real-world IoT sensing applications where data is typically collected by distributed nodes with heterogeneous data modalities, and is also rarely labeled. In this paper, we propose MMBind, a new {\emph{data binding} approach} for multimodal learning on \emph{distributed and heterogeneous IoT data}.
The key idea of MMBind is to construct a pseudo-paired multimodal dataset for model training by binding data from disparate sources and incomplete modalities through a sufficiently descriptive shared modality. 
We also propose a weighted contrastive learning approach to handle domain shifts among disparate data, coupled with an adaptive multimodal learning architecture capable of training models with heterogeneous modality combinations. 
Evaluations on ten real-world multimodal datasets highlight that MMBind outperforms state-of-the-art baselines under varying degrees of data incompleteness and domain shift, and holds promise for advancing multimodal foundation model training in IoT applications\footnote{The source code is available via https://github.com/nesl/multimodal-bind.}. 
\end{abstract}

\maketitle


\section{Introduction}

Multimodal sensing systems are increasingly deployed in real-world applications such as health monitoring \cite{ouyang2024admarker, zhang2020pdlens}, Augmented Reality (AR) \cite{chen2018marvel, apicharttrisorn2022breaking},
localization \cite{jeong2024gdtm}, and autonomous driving \cite{shi2022vips, he2023vi}. 
Most existing multimodal learning approaches require extensive training with \emph{complete multimodal data}, where each sample includes data from all sensor modalities collected simultaneously along with corresponding labels \cite{ouyang2022cosmo, liu2024focal, liu2021wavoice}. 
However, this paradigm is impractical in real-world IoT sensing applications where data is typically collected by distributed nodes. First, such data is usually \emph{heterogeneous} and \emph{incomplete}—the available multimodal data on different nodes can vary significantly, often with missing modalities or labels. For example, sensor nodes deployed in homes may contain diverse sets of modalities due to deployment constraints or privacy concerns. Sensors on a specific node may also fail dynamically, resulting in changing sensor modalities at runtime. Labeling multimodal data in real-world settings is also challenging, as data from many sensors, such as IMU, are not intuitive for human annotation \cite{hu2020fine, shuai2021millieye}.
Second, sensor data collected by distributed nodes is typically gathered \emph{at different times and locations}, resulting in disparate data that may describe similar events. For instance, in activity recognition, a wearable worn by an outdoor subject may only collect IMU data, while an indoor node may exclusively capture skeleton data of the same activity. Given the prevalence of disparate and heterogeneous data in IoT sensing, it is vital to effectively incorporate this data for multimodal learning. 





\begin{figure}
    \centering
     \setlength{\abovecaptionskip}{0.cm}
    \setlength{\belowcaptionskip}{0.cm}
    \includegraphics[width=\linewidth]{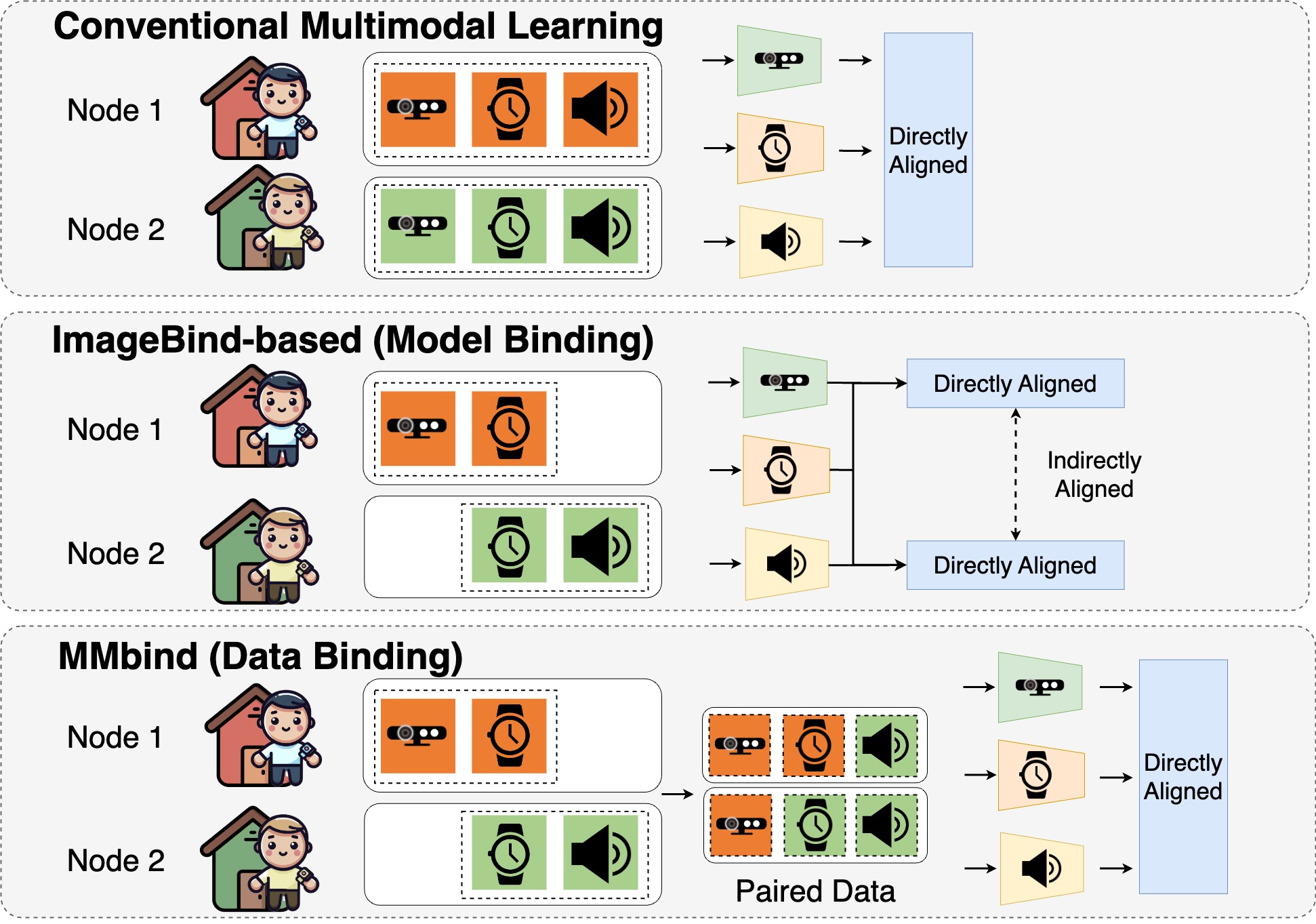}
    \caption{\name features a \emph{data binding} approach that effectively integrates disparate data with heterogeneous modalities for multimodal learning, which outperforms the \emph{model binding} approach, where various modalities are indirectly aligned through the encoder of a shared modality.
    }
    \label{fig:intro_figure}
    \vspace{-1em}
\end{figure}

Unfortunately, existing approaches struggle to address the unique challenges of multimodal learning with distributed and heterogeneous IoT data. One straightforward method is cross-modality generation, where missing modality data is generated from the remaining modalities \cite{wang2020cross}. However, training a robust generative model can be challenging in IoT sensing applications due to insufficient paired sensor data~\cite{karras2020training}.
Recently, methods like \emph{ImageBind}~\cite{girdhar2023imagebind} and others \cite{zhu2023languagebind, dai2024advancing} aim to align different modalities to the embedding space of one central modality, as shown in Figure \ref{fig:intro_figure}.
However, these model-binding approaches do not explicitly associate data modalities across different nodes and heavily depend on large amounts of data from a single central modality. Our results also indicate that they perform poorly on IoT datasets with limited data and significant domain shifts.


In this paper, we propose \name, a new framework for multimodal learning with distributed and heterogeneous IoT data. As shown in Figure \ref{fig:intro_figure}, {\name is the first \emph{data binding} approach that leverages shared modalities across distributed nodes to creating pseudo-paired multimodal data for model training. Unlike existing model-binding approaches, which do not explicitly associate data modalities across different nodes, our data-binding approach demonstrates superior performance on IoT datasets with limited training data and significant domain shifts.} Moreover, the shared modality can be either sensor data or labels, as both can effectively describe the similarity of events. 
Instead of requiring synchronized multimodal data capturing the exact same event, we show that data of different modalities observing similar events, even captured at different times and locations, can be effectively used for multimodal training. {This enables \name to construct a multimodal foundational dataset from various disparate and incomplete small datasets collected by distributed IoT nodes, which is promising to advance multimodal foundation model training in IoT applications.} 

Specifically, \name features a novel two-stage training strategy to bind distributed and heterogeneous IoT data for multimodal training, including (1) pairing incomplete data through shared modalities and (2) weighted contrastive learning with heterogeneous paired data.
In the first stage, \name gathers all data samples containing the shared modality to pre-train a unimodal encoder. When the shared modality is the data label, pre-trained large language models (LLMs) can be leveraged instead. 
\name then utilizes the feature similarity of the shared modality to match the most similar samples {from data with various domain shifts}, thereby constructing pseudo-paired multimodal data {that is associated with a corresponding similarity measurement}. \name can also be easily extended to integrate multiple incomplete datasets through successive data binding, even when utilizing different shared modalities.
In the second stage, \name employs weighted multimodal contrastive learning, {assigning varying weights to samples during contrastive learning based on the similarity of data pairs}. {This design is unique to the data binding approach, as it accommodates different contributions from pseudo-paired samples. By considering the varying quality of pseudo-paired samples due to significant domain shifts, this method enhances the robustness of the learned representations.} Furthermore, \name integrates an adaptive multimodal learning architecture capable of training models on data with heterogeneous modality combinations, leveraging both pseudo-paired and original incomplete multimodal data for model training. 
As a result, \name efficiently trains multimodal joint embeddings on large volumes of incomplete data from distributed nodes, requiring only limited or even no naturally paired multimodal data.

We extensively evaluate the performance of \name across ten different multimodal datasets {collected under uncontrolled and natural settings}, encompassing a total of nine distinct sensor modalities {and significant domain shifts across different subjects and environments}. The results indicate that \name achieves up to a 20\% accuracy improvement over several state-of-the-art baselines in both cross-node and cross-dataset data binding, and reduces training overhead on edge devices compared to training models from scratch using complete multimodal data.


In summary, we make the following key contributions:
\begin{itemize}
    \item We conduct an in-depth analysis on the impact of limited paired IoT data in multimodal learning, and show that incomplete data captured at various times and locations can be effectively used for multimodal training, as long as they describe similar events.
    \item Based on our key findings, we propose \name, {the first \emph{data binding} approach for multimodal learning with distributed and heterogeneous IoT data}. \name leverages shared modalities to bind disparate and incomplete datasets, requiring limited or even no naturally paired data.
    \item \name incorporates {a weighted contrastive learning approach based on the similarity of data pairs to effectively handle domain shifts among disparate data sources}, and an adaptive multimodal learning architecture capable of training models with heterogeneous modality combinations.
    \item Our evaluations on various datasets show that \name outperforms state-of-the-art baselines in multimodal learning, and holds promise for generating foundational datasets for multimodal learning in IoT.
\end{itemize}

\section{Related Work}
\textbf{Multimodal Sensing Systems.}
Multimodal sensing systems integrating data from multiple sensors are increasingly employed in real-world IoT applications such as speech processing ~\cite{liu2021wavoice, chen2021speech}, activity recognition~\cite{liu2023activity, huang2020activity}, and health monitoring~\cite{zhai2020health, samyuon2022health, ouyang2022cosmo}. 
Most work in this area assumes access to full-modality and time-synchronized multimodal data. 

\textbf{Distributed Sensing and Learning.}
Distributed sensor networks, consisting of spatially distributed sensing nodes, can vary from nodes in close proximity with overlapping fields of view \cite{samplawski2023heteroskedastic, grosswindhager2019snaploc}, to deployments in entirely different environments observing distinct scenes \cite{wu2022fedhome, feng2021lte}.
Training models on such spatially distributed data often employs sensor fusion algorithms \cite{jeong2024gdtm, chen2022rfcam, li2021low} that integrate information from multiple perspectives.
{Although previous multimodal federated learning studies \cite{ouyang2023harmony,wang2024towards} enable distributed model training over nodes with heterogeneous models, their upper bound is centralized unimodal learning, which does not fully align different modalities and has been shown to have inferior performance compared to \name in Section \ref{sec:evaluation}.}

\textbf{Multimodal Learning with Missing Modality.} 
When performing multimodal learning in distributed sensing systems, one must contend with \emph{heterogeneous and incomplete data}, where each sensor node captures a subset of the desired modalities. Several recent works have explored strategies for training with incomplete multimodal data. For example, \cite{liang2022expanding, huang2021learning} utilize unimodal encoders pre-trained on single-modal data to build a multimodal model, while \cite{wang2020cross, cai2018generation} generate missing modality data from available modalities. However, these methods often struggle to effectively learn cross-modal information in IoT applications due to the scarcity of high-quality multimodal pairs.
In contrast, \name leverages \emph{shared modalities} across incomplete datasets to create paired data, allowing the model to learn cross-modal correlations more efficiently.

\textbf{Model Binding Methods such as ImageBind.} 
Recent works such as \emph{ImageBind}~\cite{girdhar2023imagebind} and \emph{LanguageBind}~\cite{zhu2023languagebind} align different modalities to the embedding space of one central modality through contrastive learning. \cite{dai2024babel} expands upon this by sequentially training models of two modalities rather than relying on a central modality. All these approaches rely on \emph{model binding}, where various modalities are \emph{indirectly} aligned through transitivity, exposing the model only to (Central, X) modality pairs. In contrast, \name develops {the first \emph{data binding} approach that explicitly synthesizes pairs of (Central, X, Y) data based on the shared modalities, enabling direct multimodal learning across multiple modalities. Additionally, \name utilizes the shared modality encoder only for data pairing, while ImageBind relies upon it to perform embedding space alignment across all modalities, resulting in greater susceptibility to insufficient shared modality data.} Our evaluations in Section \ref{sec:evaluation} further show the superiority of \name on IoT datasets with limited training data and significant domain shifts.

\section{A Motivation Study}
\label{sec:motivate_incomplete}


In this section, we evaluate multimodal learning performance with limited paired data and explore the potential of leveraging disparate and incomplete data. Here, ``paired data'' refers to full-modality data, with ``pairing'' describing the process of combining separate data samples into one with complete modalities. 

\subsection{Impact of Limited Paired Data}

Obtaining full-modality and labeled multimodal data is usually impractical in real-world settings. In this section, we investigate the performance of supervised multimodal learning with different amounts of labeled paired data. 

Specifically, we evaluate using data from 15 subjects in the public RealWorld dataset \cite{sztyler2016body} to classify five human activities (walking, sitting, standing, walking upstairs, and walking downstairs). To simulate varying amounts of paired multimodal data, we divide the dataset into training sets containing 10\%, 5\%, and 1\% of labeled data, and a testing set with 90\%, while maintaining class balance. The deep learning model used consists of five CNN layers, two GRU layers, and one fully connected layer. {We repeat each experiment on five different seeds.}

\begin{figure}
     \setlength{\abovecaptionskip}{0.cm}
    \setlength{\belowcaptionskip}{-0cm}
    \centering
    \begin{minipage}{.43\linewidth}
    \setlength{\abovecaptionskip}{0.cm}
    \setlength{\belowcaptionskip}{-0.cm}
    \centering
	\includegraphics[width=\linewidth]{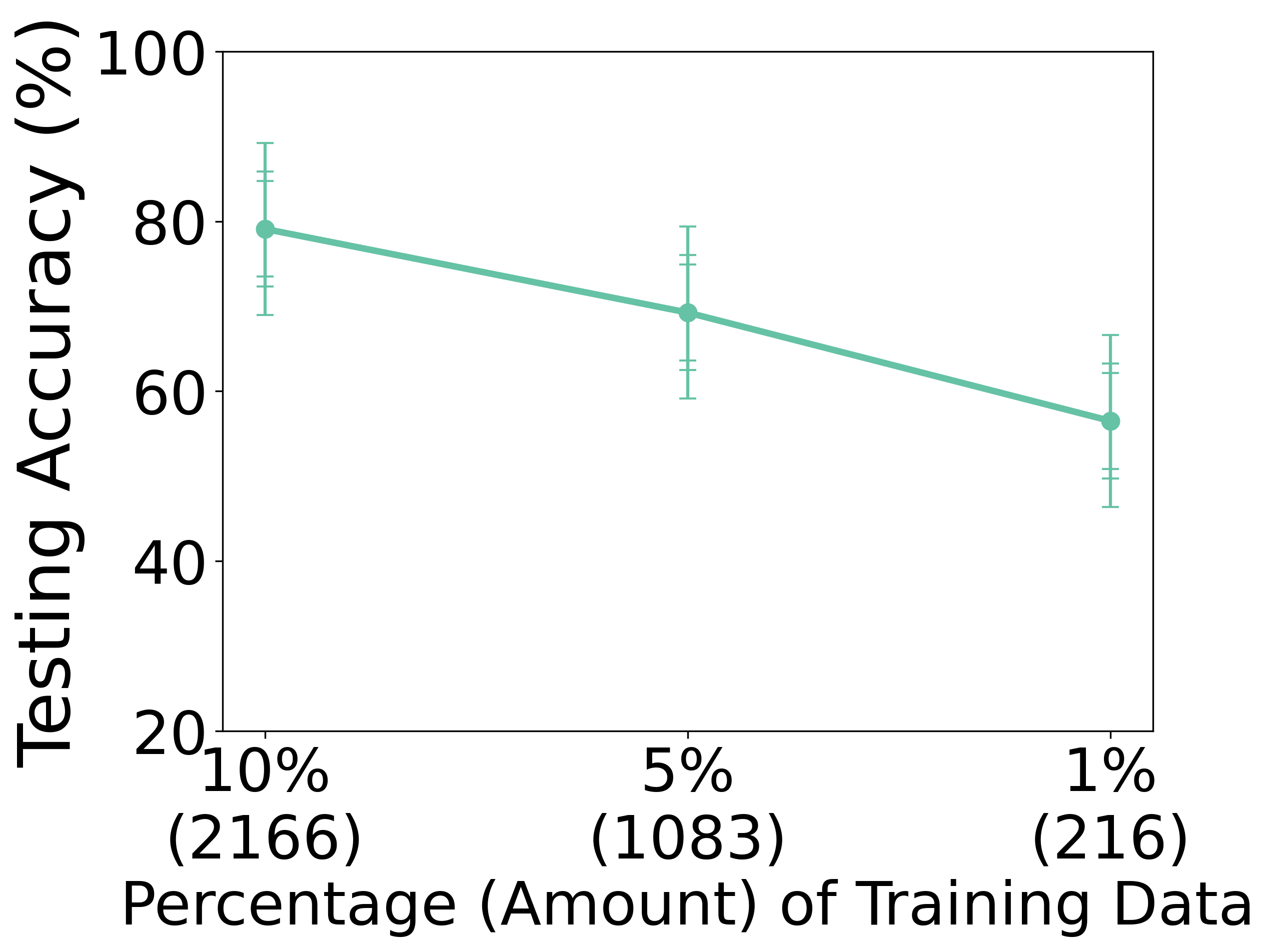}
	\caption{Impact of limited complete multimodal data.}
	\label{fig:motivation-limited-data}
    \end{minipage}
    \hspace{0.5pt}
  \begin{minipage}{.54\linewidth}
    \setlength{\abovecaptionskip}{0.cm}
    \setlength{\belowcaptionskip}{-0.cm}
    \centering
	\includegraphics[width=0.95\linewidth]{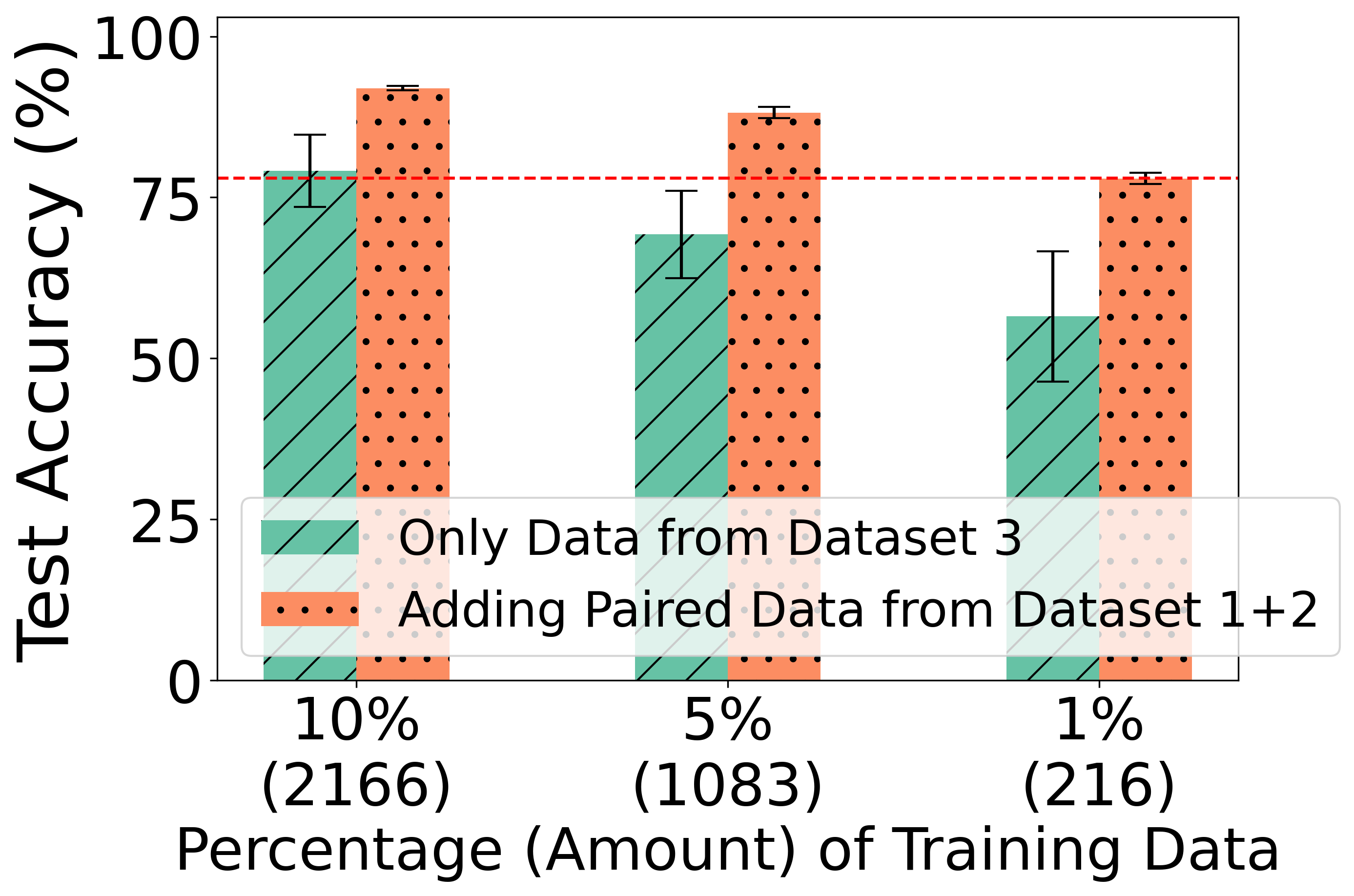}
	\caption{Adding pseudo-paired data enhances multimodal performance.
 }
	\label{fig:motivation-cross-dataset-IMU-label-bind}
  \end{minipage}
\end{figure}

Figure \ref{fig:motivation-limited-data} shows the performance of supervised multimodal learning with different amounts of labeled and paired data. Even though the task is relatively simple, the accuracy is suboptimal with limited labeled data, achieving only 55.74\% with 1\% of the training data (i.e., 216 samples). Moreover, the accuracy drops significantly when the amount of training data decreases, indicating that multimodal learning heavily relies on a sufficient amount of full-modality data. However, obtaining such full-modality data is often expensive and time-consuming in practice.  

\begin{table}
    \centering
 \setlength{\abovecaptionskip}{0.cm}
    \setlength{\belowcaptionskip}{0.cm}
    \resizebox{\linewidth}{!}{
     \begin{tabular}{c|c|c|c}
      \toprule
     Dataset  & Name & Sensor Modality & Samples \\
      \hline
    1 & MotionSense & Acc, Gyro & 12,636 \\
    2 & Shoaib & Acc, Mag & 4,500\\
    3 & RealWorld &  Acc, Gyro, Mag & 21,663 (1\%\textasciitilde10\% training)\\
      \bottomrule
    \end{tabular}}
    \caption{By binding incomplete data from Dataset 1 and 2, we can construct additional paired multimodal data for enhancing multimodal performance on Dataset 3. }
    \vspace{-1em}
    \label{table:cross_dataset}
\end{table}

\subsection{Potential of Binding Incomplete Data}

To overcome the limitation of limited paired multimodal data, we explore incorporating disparate and incomplete multimodal data. As shown in Table \ref{table:cross_dataset}, we utilize incomplete data from two additional datasets: (Acc, Gyro) data from the MotionSense dataset \cite{malekzadeh2019mobile}, and (Acc, Mag) data from the Shoaib dataset \cite{shoaib2014fusion}. These two datasets cover the same five activities as the RealWorld dataset, but were collected from entirely different subjects, devices, and environments.

Specifically, we create multimodal data pairs (i.e., Acc, Gyro, Mag) using samples from the MotionSense and Shoaib datasets. For each (Acc, Gyro) sample from MotionSense, we extract the Mag data from a randomly selected (Acc, Mag) sample from Shoaib of the same activity to form a new paired sample. The same process is applied to each (Acc, Mag) sample from the Shoaib dataset. This results in 17,136 new pseudo-paired samples. We then train a multimodal model using this paired data and fine-tune it with limited labeled full-modality data from the RealWorld dataset.

Figure \ref{fig:motivation-cross-dataset-IMU-label-bind} shows that integrating pseudo-paired data from MotionSense and Shoaib enhances performance, especially when complete multimodal data is limited. Notably, pre-training the model with pseudo-paired samples significantly boosts the performance of fine-tuning with only 1\% of training data (77.9\%), and even approaches the performance of 10\% training data (79.1\%).

This case study highlights two key insights. First, binding disparate incomplete datasets proves effective for multimodal training, which can significantly reduce the effort required for collecting and annotating multimodal data. 
Second, the effectiveness of the binding process relies on the shared information across datasets - performing binding without labels or with significant domain gaps would pose challenges.




\section{System Overview}

We now introduce \name, a new framework for multimodal learning with distributed and heterogeneous IoT data. 
The key idea is to bind disparate data to create pseudo-paired multimodal data for model training.
We first introduce applications and challenges, and then describe the problem formulation and system architecture.

\subsection{Applications and Challenges} 

\name is designed for a wide range of applications where heterogeneous sensors are deployed on distributed nodes to perform complex sensing tasks such as activity recognition \cite{ouyang2022cosmo}, scene classification \cite{zeng2021deep} or vehicle recognition \cite{shi2022vips}. 
{In these applications, \name can harness heterogeneous and incomplete data collected by distributed nodes to effectively train a multimodal neural network. \name can also construct a multimodal foundational dataset from various incomplete small datasets, which is promising to advance multimodal foundation model training in IoT applications.}

\begin{figure}
    \centering
     \setlength{\abovecaptionskip}{0.cm}
    \setlength{\belowcaptionskip}{0.cm}
    \includegraphics[width=\linewidth]{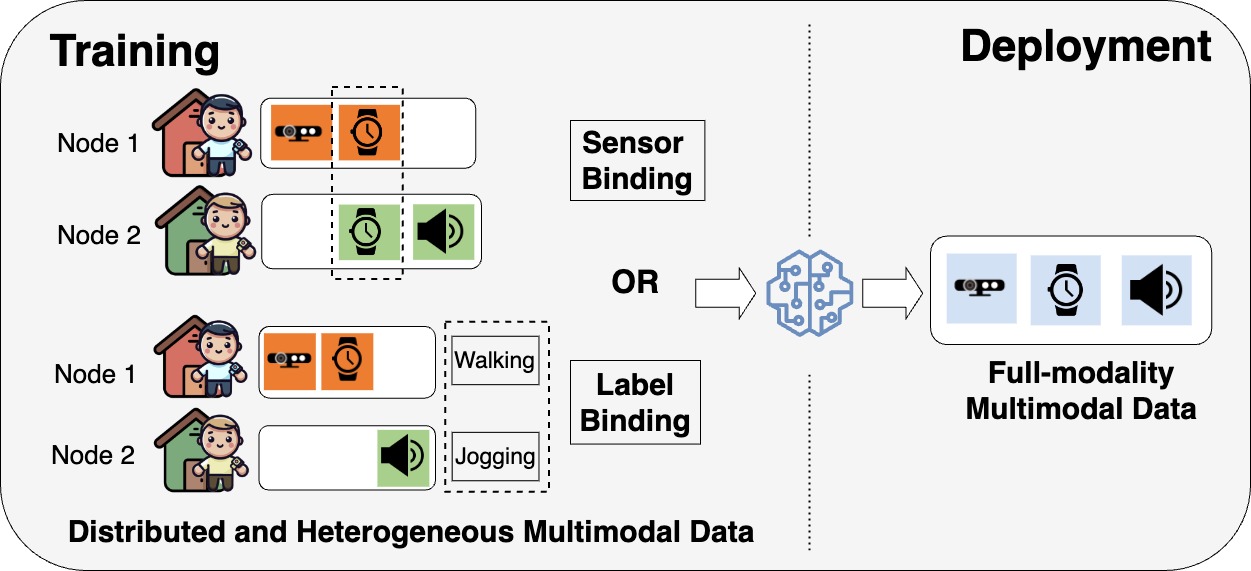}
    \caption{\name binds heterogeneous multimodal data from distributed nodes using shared sensor modalities or labels to effectively train a full-modality model.
    }
    \vspace{-1em}
    \label{fig:application_scene}
\end{figure}

We now describe two typical application scenarios. 
First, multimodal data across different nodes often exhibits significant heterogeneity, with missing modalities or labels. 
As illustrated in Figure \ref{fig:application_scene}, when the data is labeled, \name can pair samples from different nodes that describe the same or similar events. For instance, in activity recognition, one node may capture only IMU data while another records skeleton data. \name can pair IMU and skeleton data from the same or similar activity (e.g., walking and jogging) to train multimodal neural networks. 
When data labels are unavailable, \name can pair different incomplete datasets using shared sensor modalities. One node may collect IMU and audio data while another gathers IMU and skeleton data. \name can pair audio and skeleton data based on a computed similarity of their IMU data.
Second, adapting multimodal neural networks to an expanded sensing suite is also challenging. For example, consider a substantial dataset of indoor scenes collected using devices with RGB cameras and depth sensors. If a new mmWave sensor is introduced, the old dataset will become incomplete in modality.
In this scenario, \name can effectively make use of previously collected incomplete RGB and Depth data by pairing with the new mmWave data using shared modalities, reducing the amount of complete multimodal data to be recollected. In the following, we discuss these challenges in detail to motivate the design of \name.

    




\textbf{Generating data pairs with heterogeneous data.} The first challenge is how to generate effective full-modality data from incomplete data samples. First, sensor data in IoT applications is usually highly heterogeneous, with varying dimensions and data patterns, making it nontrivial to quantify the similarity of the shared modality for data pairing. Second, when multiple overlapping modalities are present, selecting the optimal common modality becomes crucial for enhancing the effectiveness of data pairing.


\textbf{Dealing with domain shift.} Another challenge is to address the varying degrees of domain gap among disparate and incomplete data. Data collected by distributed nodes can vary significantly in terms of environments or observed events, introducing substantial domain gaps. This variability increases the difficulty of accurately quantifying sample similarity and complicates the data pairing process. Furthermore, multimodal learning with improperly paired data samples can result in degraded model performance.

\subsection{Problem Formulation}
\label{sec:problem_formulation}
Without loss of generality, suppose there are two incomplete multimodal datasets collected by different nodes, i.e., $D_A$ by Node A and $D_B$ by Node B. 
The data samples in $D_A$ and $D_B$ contain up to $M$ ($M \ge 2$) different modalities in total. 
\[
\begin{aligned}
    & D_A : \{s: \mathbf{X}_A\}, \quad \mathbf{X}_A = \{ \mathbf{m}_i \mid \mathbf{m}_i \in \mathcal{M}_A \}, \\
    & D_B : \{s: \mathbf{X}_B\}, \quad \mathbf{X}_B = \{ \mathbf{m}_i \mid \mathbf{m}_i \in \mathcal{M}_B \}.
\end{aligned}
\]
Here $\mathbf{X}_k = \{ \mathbf{m}_i \mid \mathbf{m}_i \in \mathcal{M}_k \}$ ($k$ is either A or B) contains $|\mathcal{M}_k|$ different modalities, with $\mathbf{m}_i$ \textbf{denoting sensor data or data labels}. $\mathcal{M}_k \subseteq \{m_1, m_2, \ldots, m_M \}$ represents the valid data modalities in Node $k$. 

We define $D_A$ and $D_B$ as two \emph{disparate datasets} characterized by the following properties: (1) Their data is collected by different nodes at different times and locations; (2) Their data samples are incomplete in modality, i.e., $1 \le |\mathcal{M}_A| < M$, and $1 \le |\mathcal{M}_B| < M$;
(3) The samples in $D_A$ and $D_B$ share common modalities $\{\mathbf{m_s}\}$ such that $
    \mathcal{M}_A \cap \mathcal{M}_B = \{\mathbf{m_s}\} \neq \emptyset.
$
Note that the shared modality $\{\mathbf{m_s}\}$ can be either sensor data or a class label, as both can describe the similarity of events captured by distributed nodes.

\begin{figure*}
    \centering
    \setlength{\abovecaptionskip}{0.cm}
    \setlength{\belowcaptionskip}{0.cm}
    \includegraphics[width=0.9\linewidth]{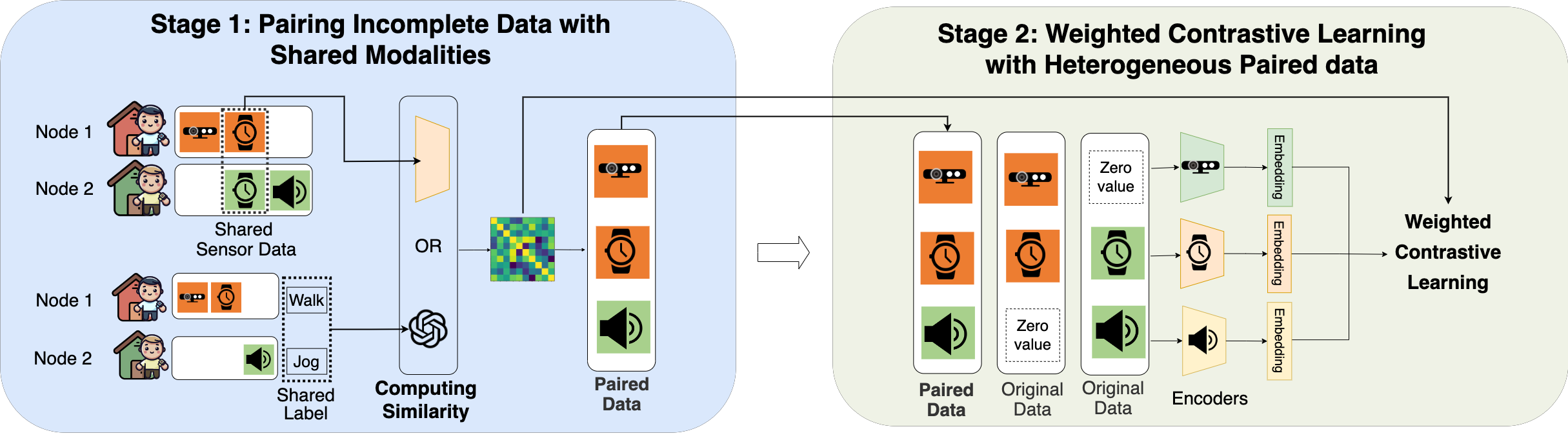}
    \caption{\name consists of two stages to bind distributed and heterogeneous IoT data for multimodal training, i.e., pairing incomplete data with shared modalities and weighted contrastive learning with heterogeneous data. 
    }
    \vspace{-1em}
    \label{fig:overview}
\end{figure*}

\name is designed to leverage incomplete multimodal samples from disparate datasets to enhance multimodal performance on data with full modalities. For example, consider $D_A: \{s: (\text{Acc}, \text{Gyro})\}$ and $D_B: \{s: (\text{Acc}, \text{Mag})\}$. \name can bind $D_A$ and $D_B$ through the similarity of shared modality $\{\mathbf{m_s} : \text{Acc}\}$, enabling  training of multimodal networks with inputs of $(\text{Gyro}, \text{Mag})$ or $(\text{Acc}, \text{Gyro}, \text{Mag})$.
In the case of label binding, given $D_A: \{s: (\text{Gyro}, \text{Label})\}$ and $D_B: \{s: (\text{Mag}, \text{Label})\}$, \name can bind $D_A$ and $D_B$ with the shared modality $\{\mathbf{m_s} : \text{Label}\}$ to train a multimodal model with inputs of $(\text{Gyro}, \text{Mag})$. We term this overall approach as \emph{data binding}, where we explicitly create full-modality samples $\{\mathbf{m}_1, \mathbf{m}_2, \mathbf{m}_s \}$, each with all $M$ modalities, to jointly train encoders $f_{enc_{m_1}}$,  $f_{enc_{m_2}}$, $f_{enc_{m_s}}$. 
In comparison, prior \emph{model binding} works such as ImageBind~\cite{girdhar2023imagebind} align encoders of different modalities to the shared modality encoder's ($f_{enc_{m_s}}$) output embedding space based on the original incomplete data samples.

\subsection{System Architecture}
The design of \name is motivated by the key insight from Section \ref{sec:motivate_incomplete} that distributed data from different modalities, captured at various times and locations, can be leveraged to enhance multimodal training if they describe similar events. Therefore, our key idea is to bind incomplete multimodal data based on the shared modality, and create pseudo-paired multimodal data samples to learn multimodal joint embeddings.
Figure \ref{fig:overview} shows the overall system architecture.

Specifically, \name features a new two-stage training strategy to bind distributed and heterogeneous IoT data for multimodal training, including pairing incomplete data with shared modalities and weighted contrastive learning with heterogeneous paired data.
In the first stage, \name gathers all samples containing the shared modality and trains the corresponding unimodal encoder. This shared modality can be either sensor data or labels, as both can describe the similarity of events captured by distributed nodes. 
When the shared modality is a data label, pre-trained large language models (LLMs) can be leveraged to measure similarity of samples. Then, \name utilizes the feature similarity of the shared modality to construct pseudo-paired multimodal data samples by matching the most similar samples across different incomplete datasets. Finally, each pseudo-paired multimodal sample will be associated with a corresponding similarity measurement.
In the second stage, \name employs an adaptive multimodal learning architecture capable of training models with heterogeneous modality combinations, by representing missing modalities with dummy inputs. Therefore, it can integrate both pseudo-paired and original incomplete multimodal data into the training process. Furthermore, to account for imperfect pairing caused by domain shifts among disparate data sources, \name introduces weighted contrastive learning on the pseudo-paired multimodal data. \name assigns varying weights to samples in contrastive learning based on the similarity of data pairs, thereby accommodating different contributions from pseudo-paired samples.
As a result, \name efficiently learns multimodal joint embeddings from distributed and heterogeneous IoT data, requiring only limited or even no naturally paired multimodal data.



\section{Design of \name}

We propose a novel two-stage framework to leverage distributed and heterogeneous IoT data for multimodal learning, namely pairing incomplete data with shared modalities and weighted multimodal contrastive learning with paired data.


\subsection{Pairing Incomplete Data with Shared Modality}
In the first stage, \name aims to pair distributed data with different modalities based on the similarity of their shared modality. We consider labels as nothing more than a natural language modality representing data in a highly compressed manner. Therefore, the shared modality can be either sensor data or labels, as both can describe the similarity of events captured by distributed nodes.


Without loss of generality, suppose $D_A : \{s: (\mathbf{m_1}, \mathbf{m_s})\}$ and $D_B : \{s: (\mathbf{m_2}, \mathbf{m_s})\}$ denotes two disparate datasets, with which we aim to train multimodal models for processing three modalities $(\mathbf{m_1}, \mathbf{m_2}, \mathbf{m_s})$.
We propose to measure the similarity between samples in $D_A$ and $D_B$ by comparing the shared modality $\{\mathbf{m_s}\}$. Since either $(\mathbf{m_1}, \mathbf{m_s})$ or $(\mathbf{m_2}, \mathbf{m_s})$ consists of samples from different modalities collected at the same time and location, the similarity across modality $\mathbf{m_s}$ indicates a strong correlation between their associated samples of modality $\mathbf{m_1}$ and $\mathbf{m_2}$.



\subsubsection{Encoder Training of the Shared Modality.}

{To enhance the effectiveness and scalability of data pairing, our key idea is to measure the similarity of \emph{feature embeddings} of the shared modality data rather than comparing high-dimensional raw sensor data. Therefore, we train a unimodal feature encoder via reconstruction to compress the shared modality data into a latent space. Such a latent space is more structured than the raw data distributions, where semantically similar samples are grouped closely together, while dissimilar samples are positioned further apart \cite{connor2021vae, kirchoff2024salsa, xurebar}.}



\begin{figure}
    \centering
     \setlength{\abovecaptionskip}{0.cm}
    \setlength{\belowcaptionskip}{0.cm}
    \includegraphics[width=\linewidth]{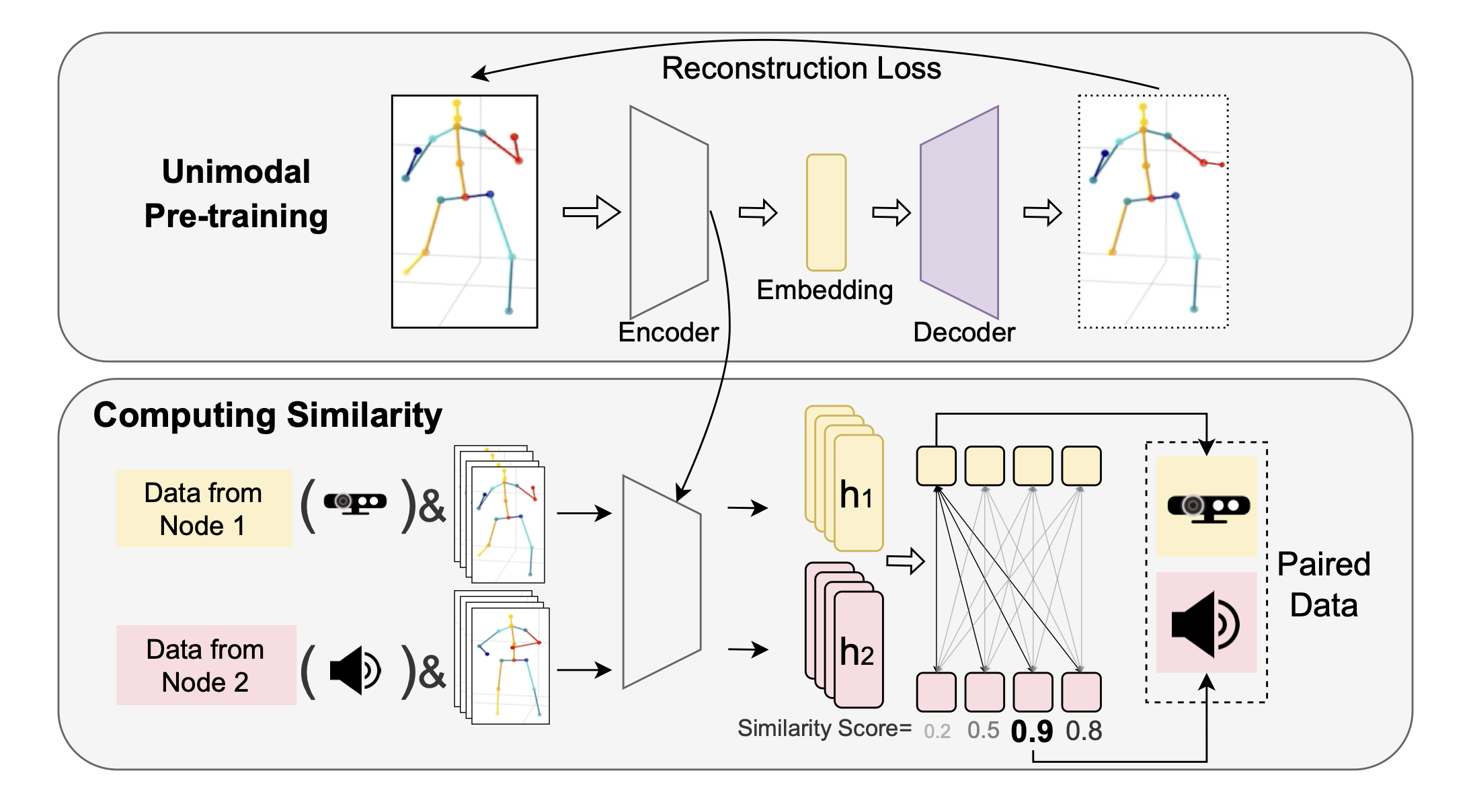}
    \caption{Pairing incomplete data with the shared modality. We train an encoder on the shared modality data through self-supervised reconstruction loss, and then use the feature similarity of shared modality data to pair remaining modalities from different datasets. 
    }
    \vspace{-1em}
    \label{fig:measure_sim}
\end{figure}

As shown in Figure~\ref{fig:measure_sim}, \name first takes the union of all data from the shared modality, i.e., $\mathbf{m_s}$ from $D_A$ and $D_B$, to train a unimodal encoder of $\mathbf{m_s}$. Specifically, we train the unimodal encoder in a self-supervised manner using an autoencoder. The autoencoder consists of an \emph{encoder} $f_{enc_s}(\cdot)$ that compresses the data into a smaller latent space, followed by a \emph{decoder} $f_{dec_s}(\cdot)$ that reconstructs the original input data from the latent embedding. The reconstruction loss is defined as the L2 distance between the decoder's output and the original input data. Suppose the reconstructed data is $\mathbf{m'}_s = f_{dec_s}(f_{enc_s}(\mathbf{m_s}))$, then the reconstruction loss is given by:
\begin{equation}
    \mathcal{L}_{reconstruct} = || \mathbf{m'}^j_s - \mathbf{m}^j_s ||^2, j=1, ..., |D_A|+|D_B|  \label{eq:reconstruct_loss}
\end{equation}
Through the training process, the encoder learns to compress semantically essential features of the input data into a smaller latent space, while effectively filtering out irrelevant features. 




Similarly, when \emph{labels} serve as the shared modality, we can directly utilize pre-trained \emph{large language models} (LLMs) to measure the semantic similarity of natural language labels, eliminating the need to train unimodal encoders. {
For example, we employ an encoder-only LLM, paraphrase-MiniLM-L6-v2, trained by Sentence-BERT \cite{reimers2019sentence}, to measure the similarity of language labels. }
In this paper, we primarily focus on cases where labels sufficiently describe the similarity of the data.
However, class labels can sometimes fail to fully capture the similarity of an event, leading to incongruent data associations. For example, sensor data labeled identically but collected from different locations (e.g., worn inwards versus outwards) may exhibit significant differences. In such cases, we can enhance the labels by incorporating additional metadata in structured formats, such as the subjects' gender, environment, and devices used, alongside the class labels. 
The LLM then identifies the most semantically similar samples and calculates their corresponding similarities based on the language tokens of the labels.

\subsubsection{Generating Pseudo-Paired Data}
After unimodal pre-training, the encoder of the shared modality serves as a benchmark for measuring the similarity between datasets \( D_A \) and \( D_B \). \name then utilizes the feature similarity of the shared modality to pair these datasets. Specifically, the data samples of shared modality from \( D_A \) and \( D_B \) are fed into the feature encoder of \( \mathbf{m_s} \). We have:
\begin{align*}
     & \mathbf{h}^j_{s} = f_{enc_s}(\mathbf{m}^j_{s}), j=1, ..., |D_A|. \\
     & \mathbf{h}^k_s = f_{enc_s}(\mathbf{m}^k_s), k=1, ..., |D_B|.
\end{align*}
As mentioned earlier, when \emph{labels} serve as the shared modality, we utilize existing pre-trained LLMs to tokenize the language labels and derive their corresponding embeddings $\mathbf{h}^j_s$ and $\mathbf{h}^k_s$.
Next, we compute the pairwise cosine similarity $a_{jk}$ between features of the shared modality \( \mathbf{h}^j_{s} \) from \( D_A \) and \( \mathbf{h}^k_{s} \) from \( D_B \), resulting in a similarity matrix \( \mathbf{A} \in \mathbb{R}^{|D_A| \times |D_B|} \). {We use pairwise cosine similarity as the pairing metric, as it is a common method for measuring the similarity of feature representations in deep learning models. 
However, other similarity measurement metrics, such as Euclidean distance, can be also employed in our design.}



Then, \name constructs pseudo-paired multimodal data samples by matching the most similar samples between disparate datasets. For example, for the \( j \)-th data sample in \( D_A \), denoted as \( \mathbf{X}^j_A = (\mathbf{m}^j_1, \mathbf{m}^j_s) \), we search \( D_B \) for the sample with the highest feature similarity in the shared modality \( \mathbf{m}_s \), i.e.,
\begin{equation}
p_j = \arg\max_{1 \leq k \leq |D_B|} a_{jk}.
\end{equation}
Here, \( \arg\max\) denotes the index where \( a_{jk} \) achieves its maximum value within \( \mathbf{a}_j \). Therefore, the corresponding paired sample from $D_B$ will be $\mathbf{X}^{p_j}_B = (\mathbf{m}^{p_j}_2, \mathbf{m}^{p_j}_s)$. Moreover, each pseudo-paired multimodal data sample will also be associated with the corresponding similarity measurement $a_{jp_j}$. This process occurs for every $j_{th}$ data sample in $D_A$ and for every $k_{th}$ data sample in $D_B$:
\begin{align*}
    & \mathbf{X}^j_{Ap} = (\mathbf{m}^j_1, \mathbf{m}^j_s, \mathbf{m}^{p_j}_2, a_{j{p_j}}), j=1, ..., |D_A|\\
    & \mathbf{X}^k_{Bp} = (\mathbf{m}^{p_k}_1, \mathbf{m}^k_s, \mathbf{m}^{k}_2, a_{{p_k}k}), k=1, ..., |D_B|
\end{align*}
As the pairing process is repeated \emph{for every sample of the two datasets}, the combined pseudo-paired dataset $D_P$ will have $|D_A| + |D_B|$ data samples in total, i.e., 
\begin{equation*}
    D_P: \{s: \mathbf{X}_P\}, \quad \mathbf{X}_P = \{ \mathbf{m}_1, \mathbf{m}_2, \mathbf{m}_s, a\}
\end{equation*}

{When pairing large datasets or when feature embeddings of shared modalities have high dimensionality, pairwise binding by exhausting the cosine similarity between every pair of samples can be highly time-consuming. To address this, we could explore group-based pairing to reduce the number of data samples in search of optimal pairs, or employ techniques such as Principal Component Analysis (PCA) \cite{abdi2010principal} to reduce the dimensionality of features, further accelerating the data pairing process.}

\subsubsection{Pairing Considerations} We now discuss certain considerations before performing data pairing in \name.

\textbf{Selection of the binding modality.} 
\label{sec:select_common_modality}
{The quality of the pseudo-paired dataset depends heavily on whether features of the shared modality can accurately measure the semantic similarity of different data samples. As a result, the performance of MMBind heavily relies on the shared modalities.} Therefore, when the disparate datasets $D_A$ and $D_B$ have \emph{multiple} overlapping modalities, it is crucial to carefully select the shared modality \(\mathbf{m}_s\) for data binding.

Here, we summarize two major properties that a good shared modality should possess for data pairing in \name. {If the modality does not exhibit such properties, there may be no performance benefits when binding different datasets, as shown in Section \ref{sec:exp_select_common_modality}. }First, the shared modality itself should be able to effectively \emph{distinguish different events in the downstream task}. Otherwise, the features generated by the unimodal encoder will fail to differentiate between similar and dissimilar events.
For example, in the task of classifying an object's color, mmWave radar is not an appropriate choice for a common modality, as the similarity measured in the latent space would not reflect the object's color. Using this mismatched multimodal data for training is likely to degrade the final performance. 

Second, the shared modality should \emph{effectively generalize across different domains}. Data collected by distributed nodes at different times and locations often exhibits significant domain shifts. If the shared modality data is not robust to changing scenarios, the domain gap between disparate datasets will overshadow the similarity of the same events.
For example, in activity recognition, skeleton data remains robust across different subjects and environments, making it suitable for data pairing. In contrast, WiFi data is highly sensitive to environmental changes, leading to significant degradation in pairing performance.
We explore the choice of the common modality and its impact in Section \ref{sec:exp_select_common_modality}.

\textbf{Binding more than two datasets.} The above approach can be easily extended to integrate multiple incomplete datasets through successive data binding. {When there are multiple datasets, the shared modality does not need to be the same across different datasets, as evaluated in Section \ref{sec:exp_multiple_datasets}.} However, while incorporating pseudo-paired data typically enhances performance compared to using no pairings, the improvement does not always correlate with the number of paired samples. When integrating multiple datasets, it is essential to consider their heterogeneity. For instance, there may be substantial differences in data distribution among the paired datasets due to reasons such as variations in sensor placement.
We also evaluate and discuss the impact of additional paired data in Section \ref{sec:evaluation-cross-dataset}.

  


\subsection{Weighted Contrastive Learning with Heterogeneous Paired Data}

In the second stage, our goal is to train the feature encoders to learn multimodal joint representations by leveraging both the pseudo-paired data and the original incomplete data. 


\subsubsection{Adaptive Multimodal Learning with Heterogeneous Modality Combinations} \name employs an adaptive multimodal learning architecture that is capable of training models with heterogeneous modality combinations. As shown in Figure~\ref{fig:overview}, we aim to train a multimodal neural network containing encoders of all modalities. \name leverages both pseudo-paired multimodal samples and original incomplete multimodal samples for model training. The aggregated training dataset contains samples with modalities $(\mathbf{m}_1, \mathbf{m}_s)$, $(\mathbf{m}_s, \mathbf{m}_2)$, and $(\mathbf{m}_1, \mathbf{m}_2, \mathbf{m}_s)$. Samples with a missing modality will include a zero-valued dummy representation such that our training dataset is:
\begin{equation*}
    D_S : \{s: \Tilde{\mathbf{X}}_A,
    \Tilde{\mathbf{X}}_B, \Tilde{\mathbf{X}}_P\},
\end{equation*}
where $\Tilde{\mathbf{X}}_A = (\mathbf{m}_1, \mathbf{0}, \mathbf{m}_s)$, $\Tilde{\mathbf{X}}_B = (\mathbf{0}, \mathbf{m}_2, \mathbf{m}_s)$, $\Tilde{\mathbf{X}}_P = (\mathbf{m_1}, \mathbf{m}_2, \mathbf{m}_s)$.

The advantages of such a multimodal learning architecture are as follows. First, it allows us to fully utilize both original incomplete and pseudo-paired samples, thereby increasing the amount of training data. Although the original datasets $D_A$ and $D_B$ contain only incomplete samples, they are naturally paired and exhibit more ``accurate'' information for cross-modality alignment between $\mathbf{m}_s$ and the remaining modalities. In contrast, while the pseudo-paired dataset $D_P$ may include some noisy, mis-paired samples, it facilitates better alignment across the paired modalities $\mathbf{m}_1$ and $\mathbf{m}_2$. Second, by increasing the diversity of modality combinations in the training data, the full-modality multimodal model will be more robust to various missing modalities during inference. We also show the deployment robustness of \name on different modality combinations in Section \ref{sec:deployment_robustness}.





\begin{figure}
    \centering
     \setlength{\abovecaptionskip}{0.cm}
    \setlength{\belowcaptionskip}{0.cm}
    \includegraphics[width=\linewidth]{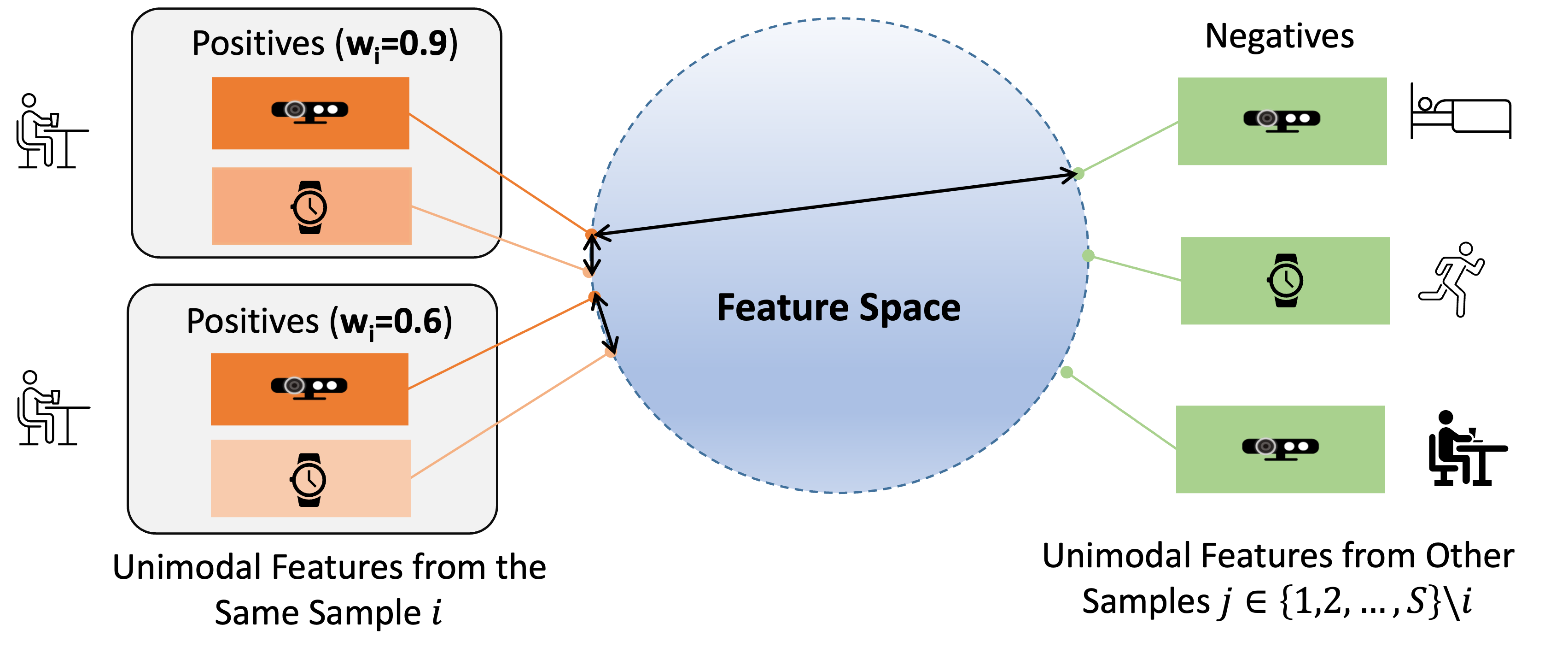}
    \caption{Weighted contrastive learning with pairing similarity. We weigh the contrastive loss of each sample based on the similarity of data pairs to accommodate different similarities of pseudo-paired samples.}
    \vspace{-1em}
    \label{fig:weighted_contarstive}
\end{figure}

\subsubsection{Weighted Contrastive Learning based on Pairing Similarity}
Leveraging both pseudo-paired and original incomplete data, our goal is to train feature encoders for different modalities, ensuring that their output embeddings are aligned within a unified feature space.
Additionally, it is also crucial to account for diverse levels of similarity among the pseudo-paired samples that arose during the first stage of data pairing.

To achieve this, we perform contrastive learning on the unimodal features of different modalities. As shown in Figure \ref{fig:weighted_contarstive}, this will push unimodal features within the same multimodal sample closer together in the feature space, while separating unimodal features from different samples. To accommodate different output embedding sizes from different encoders, we use multi-layer perceptrons to map the features to the same dimension. After applying a normalization to ensure they lie on the unit hyperplane, we have embeddings $\mathbf{z}_q \in \mathbb{R}^{F}, q= 1,...M$, with $F=128$ in our experiments.


Furthermore, to effectively account for the diverse similarity of paired samples, \name performs weighted contrastive learning on pseudo-paired multimodal data. In previous contrastive learning approaches, the positive samples originate from naturally-paired multimodal samples corresponding to the same event. However, the data pairing of \name relies on the feature similarity of the shared modality, which can lead to subpar data pairs, especially when significant domain gaps exist between disparate datasets.

To address this challenge, we propose to incorporate the degree of similarity among the positive pairs into contrastive learning. Specifically, we introduce weights to adjust the contrastive learning loss of each sample based on the similarity of data pairs, thereby accommodating different contributions of various pseudo-paired samples. The loss function for the weighted multimodal contrastive loss is:
\begin{equation}
\mathcal{L}_{\mathrm{w\_contrast}} = - \sum_{i \in \mathcal{S}} w_{i} \sum_{1\le p, q \le M} \log \frac{\exp(\mathbf{z}^i_q \cdot \mathbf{z}^i_p / \tau)}{\sum_{j\neq i} \exp(\mathbf{z}^i_p \cdot \mathbf{z}^j_q / \tau)}
\end{equation}
where \((\mathbf{z}^i_q, \mathbf{z}^i_p)\) denotes the positive samples with unimodal features of modality $p$ and $q$ from the same paired sample $\mathbf{X}^i$, while \((\mathbf{z}^i_q, \mathbf{z}^j_p)\) denotes the negative samples with unimodal features from different samples $\mathbf{X}^i$ and $\mathbf{X}^j$. Moreover, \(w_{i}\) represents the weight assigned to the $i_{th}$ paired data sample \( \mathbf{X}^i = (\mathbf{m}^i_1, ..., \mathbf{m}^i_M)\), which is equal to the normalized pairing similarity $a_{i}$. $\mathcal{S}$ denotes the set of indices for data samples in the aggregated dataset $D_S$.   
Therefore, through weighted multimodal contrastive learning, the model is trained to learn multimodal joint embeddings while emphasizing data samples with higher similarity. This approach helps mitigate the impact of mis-paired samples, improving overall model performance.

\section{Evaluation}
\label{sec:evaluation}

\subsection{Methodology}

\subsubsection{Datasets} 
We evaluate the performance of \name using ten real-world multimodal datasets with various data modalities and domain shifts. Table \ref{table:summary_dataset} summarizes the datasets and their data splitting. We use the UTD~\cite{utd-dataset}, MM-Fi~\cite{yang2024mm}, PAMAP2~\cite{reiss2012introducing}, and SUN-RGBD ~\cite{song2015sun, silberman2012indoor, janoch2013category, xiao2013sun3d} (subset) datasets to simulate cross-node (subject) data pairing. We use MotionSense~\cite{malekzadeh2019mobile}, Shoaib~\cite{shoaib2014fusion} and RealWorld~\cite{sztyler2016body} datasets, and GR4DHCI~\cite{wang2023multimodal}, DHG~\cite{devineau2018deep} and Briareo~\cite{manganaro2019hand} dataset to evaluate the performance of \name in cross-dataset data binding. When performing cross-node data binding, we construct two \emph{unlabeled sub-datasets} $\{\mathbf{m}_1, \mathbf{m}_s\}$ and $\{\mathbf{m}_2, \mathbf{m}_s\}$, where $\mathbf{m}_s$ can be either a shared sensor modality or the data label. A small \emph{modal-complete, labeled} dataset $\{\mathbf{m}_1, \mathbf{m}_2\}$ is used for supervised finetuning, while a larger dataset with the same modalities is used for testing. For sensor data as binding modality, we use \emph{Acc} as the shared modality for \textbf{UTD-MHAD}, \textbf{PAMAP2}, \emph{Skeleton} for \textbf{MM-Fi}, and \emph{Image} for \textbf{SUN-RGBD}. Details on cross-dataset binding are in Section \ref{sec:evaluation-cross-dataset}.  
{Among these datasets, PAMAP2, SUN-RGBD, RealWorld, MotionSense, GR4DHCI, and Briareo are collected under uncontrolled, natural, and real-world settings, exhibiting significant domain shifts across different subjects and environments.}

\begin{table}
    \centering
    \resizebox{\linewidth}{!}{
     \begin{tabular}{c|ccccc}
      \thickhline
      Setting & Dataset & Modality & \makecell{Nodes\\(Subjects)} & Class  & Sample  \\
      \midrule
     \multirow{3}{*}{\makecell{Cross \\ Node \\ (Intra\\ Dataset)}} &  UTD & \begin{tabular}[c]{@{}c@{}}Acc, Gyro, Skeleton\end{tabular} & 4/2/2  & 27 & 864 \\
    & MMFi & \begin{tabular}[c]{@{}c@{}}Depth, Radar,\\ Skeleton, WiFi\end{tabular} & 20/10/20 & 27 & 1,080 \\
    & PAMAP2 & Acc, Gyro, Mag & 4/2/2 & 30 & 9,611 \\
    & SUN-RGBD & Image, Depth, SemSeg & N/A & 5 & 4,620 \\
       \hline
   \multirow{3}{*}{ \makecell{Cross \\ Dataset \\(Activity)}} &  MotionSense & Acc, Gyro & 24 & 6 & 12,636 \\
     &  Shoaib-right & Acc, Mag & 10 & 7 & 4,500 \\
     &  Shoaib-left & Acc, Mag & 10 & 7 & 4,500 \\
     &  Shoaib-wrist & Acc, Mag & 10 & 7 & 4,500 \\
   &  RealWorld & Acc, Gyro, Mag & 15 & 8 & 21,663 \\
   \hline
   \multirow{3}{*}{\makecell{Cross \\ Dataset \\ (Gesture)}} &  GR4DHCI & Skeleton, IR & 16 & 7 & 7,339 \\
     &  DHG & Skeleton, Depth & 20 & 14 & 2,800 \\
   &  Briareo & Skeleton, Depth, IR & 40 & 12 & 1,440\\
      \thickhline
    \end{tabular}}
    \caption{Summary of the multimodal datasets. The cross-node setting shows the number of nodes/subjects in the incomplete pre-training, fine-tuning, and testing datasets in the format X/Y/Z, respectively.
    }
    \vspace{-2em}
    \label{table:summary_dataset}
\end{table}





\subsubsection{Baselines} The baseline approaches are as follows.

\noindent \textbf{Unimodal Learning (Unimodal).} This approach separately trains unimodal encoders using isolated data from different incomplete datasets, employing either autoencoder-based unsupervised learning (shared modality $\mathbf{m}_s$ is sensor data) or supervised learning ($\mathbf{m}_s$ as data labels). The trained unimodal encoders are then fine-tuned using limited labeled and paired multimodal data.



\noindent \textbf{Multimodal Learning with Incomplete Modality (MIM) \cite{parthasarathy2020zero}.} This approach directly trains multimodal models on the incomplete multimodal data, by representing missing modality data with a zero-valued input. 



\noindent \textbf{Modality Prompting for Missing Modality (MPM )\cite{lee2023multimodal}.} This approach builds on the above MIM method by incorporating a prompt to inform the neural network about the presence of input modalities. For example, the prompt vector (1, 0, 1) signifies data samples with modalities $(\mathbf{m}_1, \mathbf{0}, \mathbf{m}_s)$.


\noindent \textbf{Cross-Modality Generation (CMG) \cite{wang2020cross}.} This approach trains a cross-modal generative model between the shared modality and every other modality, enabling data generation for missing modalities. This method is applicable only when the shared modality is sensor data.


\noindent \textbf{Dual Contrastive Multimodal Learning (DCM) \cite{ma2019unpaired}.}  
Building on the above MIM approach, this method calculates contrastive loss using only the embeddings corresponding to the actual data, rather than all modality embeddings. It is applicable only when the shared modality is sensor data.

\noindent \textbf{ImageBind \cite{girdhar2023imagebind}.} We utilize ImageBind's approach of loading a pre-trained encoder of the shared modality, and utilizing its frozen representations to align different modalities to a shared embedding space. Note that this is similar to the previous DCM baseline except from loading and freezing the common modality's encoder. 


\noindent \textbf{Lower Bound.} 
This approach trains a full-modality neural model from scratch on the labeled finetuning dataset, establishing the lower bound performance of \name\ by ignoring incomplete data samples during training.

\noindent \textbf{Theoretical Upper Bound.} This method  pre-trains a full-modality model using the same amount of \emph{naturally paired data} as the incomplete samples in \name, after which it is subsequently fine-tuned. This should theoretically yield optimal performance due to the use of naturally paired data.

\subsubsection{Configurations} For the unimodal feature encoders, we use CNN+RNN layers or transformers to extract deep features. Specifically, 2D-CNN is applied to inertial data and image-like data, 3D-CNN to skeleton data, transformers to radar data, and multi-layer perceptrons are used for the classifier. We employ an encoder-only LLM, paraphrase-MiniLM-L6-v2 trained by Sentence-BERT \cite{reimers2019sentence}, to measure the similarity of language labels. {We repeat each experiment on five different seeds.} 



\begin{figure}
\centering
     \setlength{\abovecaptionskip}{0.cm}
    \setlength{\belowcaptionskip}{0.cm}
        \includegraphics[width=\linewidth]{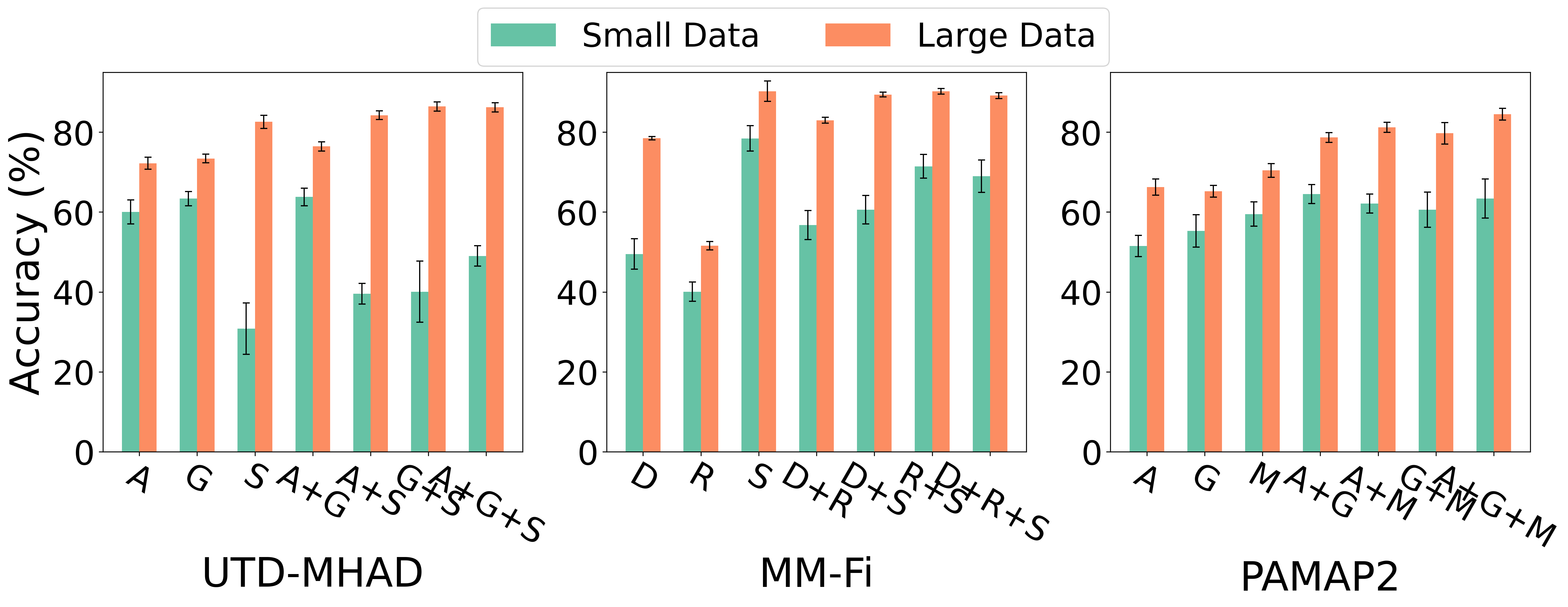}
	\caption{Performance of different modality combinations. A: Accelerometer, G: Gyroscope, S: Skeleton, D: Depth, R: mmWave Radar, M: Magnetic.
    }
    \vspace{-1em}
	\label{fig:all-modality}
\end{figure}

\subsection{Performance on Different Datasets}

\subsubsection{Understanding different multimodal datasets} We first analyze the characteristics of different datasets through comparing supervised learning performance on various modality combinations. Figure~\ref{fig:all-modality} shows results from the UTD-MHAD, MM-Fi, and PAMAP2 datasets across scenarios with limited (Small Data) and ample (Large Data) training samples. When training data is limited, integrating additional modalities often leads to overfitting, diminishing performance compared to single-modal approaches. However, with sufficient training data, multimodal learning outperforms single-modal methods, motivating the design of \name to generate additional paired multimodal samples for enhancing multimodal performance.
Moreover, with enough training data, certain single sensor modalities can achieve notably good performance on their own. For example, skeleton data in the MM-Fi dataset achieves about 90\% accuracy, indicating its effectiveness in distinguishing different events and potential as a shared modality for binding disparate datasets. 




\begin{table}
\resizebox{\columnwidth}{!}
{%
\begin{tabular}{c|cc|cc|cc|cc}
\thickhline
\multirow{2}{*}{Datasets}  & \multicolumn{2}{c|}{UTD-MHAD}  & \multicolumn{2}{c|}{MM-FI} & \multicolumn{2}{c|}{PAMAP2} & \multicolumn{2}{c}{SUN-RGBD} \\ 
 & Acc & F1 & Acc & F1 & Acc & F1 & Acc & F1  \\ \hline 
Lower Bound & 40.41           & 0.380           & 65.74           & 0.654           & 64.51           & 0.609 & 34.80 & 0.331           \\
Unimodal    & \underline{69.04} & \underline{0.646} & 53.91           & 0.532           & 59.44           & 0.528 & 33.23 & 0.323          \\
MIM         & 62.23           & 0.590            & 68.31           & 0.676           & 63.38           & 0.567 
    & 40.33 & 0.390
\\
MPM         & 69.74           & 0.666           & 70.71           & 0.701           & 64.15           & 0.592  
    & 41.10 & 0.394
\\
CMG         & 61.69           & 0.592           & \underline{72.17} & \underline{0.722} & 61.62           & 0.577  & 35.96 & 0.342          \\
DCM         & 59.25           & 0.563           & 68.26           & 0.678           & \underline{64.43} & \underline{0.597} & 40.14 & 0.374\\
ImageBind         & 62.99           & 0.612           & 70.04           & 0.698           & 61.58 & 0.568 
& \underline{44.08} & \underline{0.418}\\
\textbf{MMBind}   & \textbf{78.86} & \textbf{0.763} & \textbf{77.72} & \textbf{0.775} & \textbf{69.08} & \textbf{0.654} & \textbf{44.56} & \textbf{0.419}\\ \hline
Upper Bound & 78.68           & 0.768           & 72.45           & 0.720           & 68.87           & 0.636  
    & 42.61 & 0.398
\\
\thickhline
\end{tabular}
}
\caption{Performance on cross-node data binding with a shared sensor modality. 
}
\vspace{-2em}
\label{table:cross_subject_sensor_bind}
\end{table}

\begin{table}
\resizebox{\columnwidth}{!}
{%
\begin{tabular}{c|cc|cc|cc|cc}
\thickhline
\multirow{2}{*}{Datasets}  & \multicolumn{2}{c|}{UTD-MHAD}  & \multicolumn{2}{c|}{MM-FI} & \multicolumn{2}{c|}{PAMAP2} & \multicolumn{2}{c}{SUN-RGBD} \\ 
 & Acc & F1 & Acc & F1 & Acc & F1 & Acc & F1\\ \hline 
Lower Bound  & 40.09 & 0.378  & 65.74 & 0.654  & 63.52 &	0.581 & 30.420 & 0.290 \\ 
Unimodal  & 69.54 &	0.668  & 44.35 & 0.432 & 67.68 &	0.637 & 32.68 & 0.247\\ 
MIM & \underline{75.84} &	\underline{0.729}  & 61.80 & 0.611 & 68.45 &	0.657 & \underline{53.52} & \underline{0.533}\\ 
MPM & 73.95 &	0.726   & \underline{63.44} & \underline{0.623} & \underline{69.23} &	\underline{0.678} & 52.18 & 0.506\\ 
\textbf{MMBind} & \textbf{83.04} & \textbf{0.813} & \textbf{77.09} &	\textbf{0.775} & \textbf{74.44} & \textbf{0.723} & \textbf{55.54} & \textbf{0.549}\\ \hline 
Upper Bound  & 85.63 &	0.844 & 84.12	 & 0.841  & 73.65 & 0.722 & 63.91 & 0.637\\ 
\thickhline
\end{tabular}}
\caption{Performance on cross-node data binding with shared data labels. 
}
\vspace{-2em}
\label{table:cross_subject_label_bind}
\end{table}

\subsubsection{Sensor data as the binding modality} 
We then evaluate the performance of \name when using sensor data to pair different incomplete datasets for training. As shown in Table \ref{table:cross_subject_sensor_bind}. \name consistently outperforms all other baselines. For example, on the UTD-MHAD dataset, \name achieves a mean accuracy improvement of 9.82\% and an F1 score improvement of 11.7\% over the best baseline. The lower accuracy of all methods on SUN-RBGD can be attributed to the challenging nature of scene classification from limited high-dimensional data.
Moreover, the performance of \name approaches or even surpasses the theoretical upper bound achieved with naturally paired data. We attribute this result to the nature of contrastive pre-training, where training with naturally paired samples may inadvertently capture irrelevant similarities across subjects and environments. In contrast, \name pairs samples from incomplete datasets based on task-relevant features, thereby mitigating the risk of learning confounding features.

Additionally, \name outperforms the ImageBind model-binding approach across all datasets, with the exception of comparable performance on SUN-RGBD. We show that with limited data in IoT, the explicit \emph{data pairing} approach achieves superior performance by directly applying contrastive learning across all the modalities, rather than relying on transitivity to implicitly align embedding spaces in model binding.



\subsubsection{Label as the binding modality} Table \ref{table:cross_subject_label_bind} shows the performance of \name and various baselines when using labels to bind different incomplete data samples. The results show that \name consistently outperforms other baselines by pairing unimodal data of different modalities using labels. For example, on the MM-Fi dataset, \name outperforms the best baseline by 13.65\% in mean accuracy and 15.2\% in F1 score. Moreover, the performance of \name in label binding meets or exceeds that of using sensor data as the shared modality, suggesting that data labels are generally more effective in assessing the similarity of data samples.


{
\subsection{Dealing with Domain Shift}
\label{sec:evaluation-cross-dataset}
In this section, we evaluate the performance of \name under substantial domain shift.

\begin{figure}
     \setlength{\abovecaptionskip}{0.cm}
    \setlength{\belowcaptionskip}{0.cm}
    \centering
  \begin{subfigure}{.49\linewidth}
    \setlength{\abovecaptionskip}{0.cm}
    \setlength{\belowcaptionskip}{-0.cm}
    \centering
     \includegraphics[width = \textwidth]{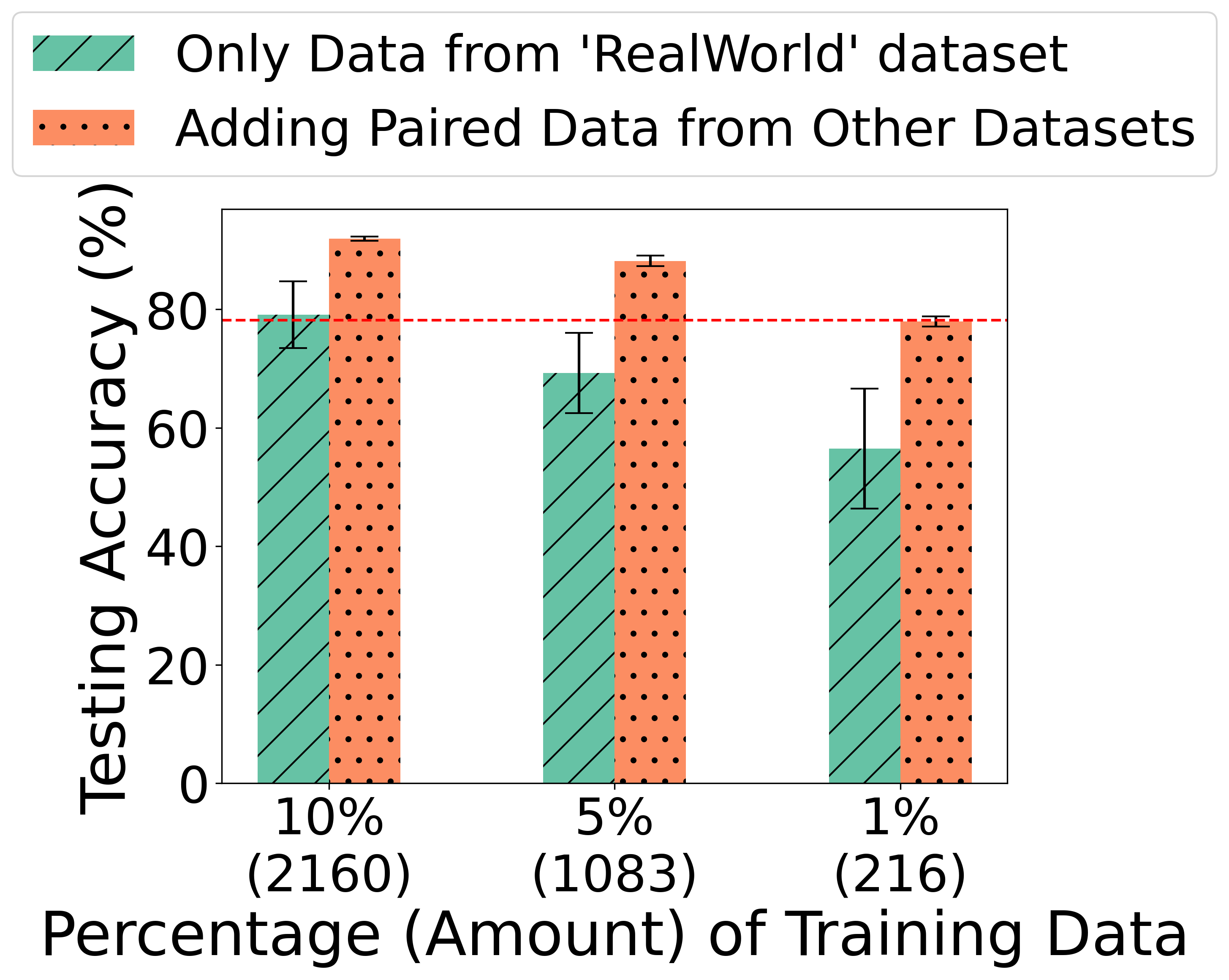}
    \caption{Label binding.}
      \label{fig:cross-dataset-IMU-label-bind}
  \end{subfigure}
      \begin{subfigure}{.49\linewidth}
    \setlength{\abovecaptionskip}{0.cm}
    \setlength{\belowcaptionskip}{-0.cm}
    \centering
     \includegraphics[width = \textwidth]{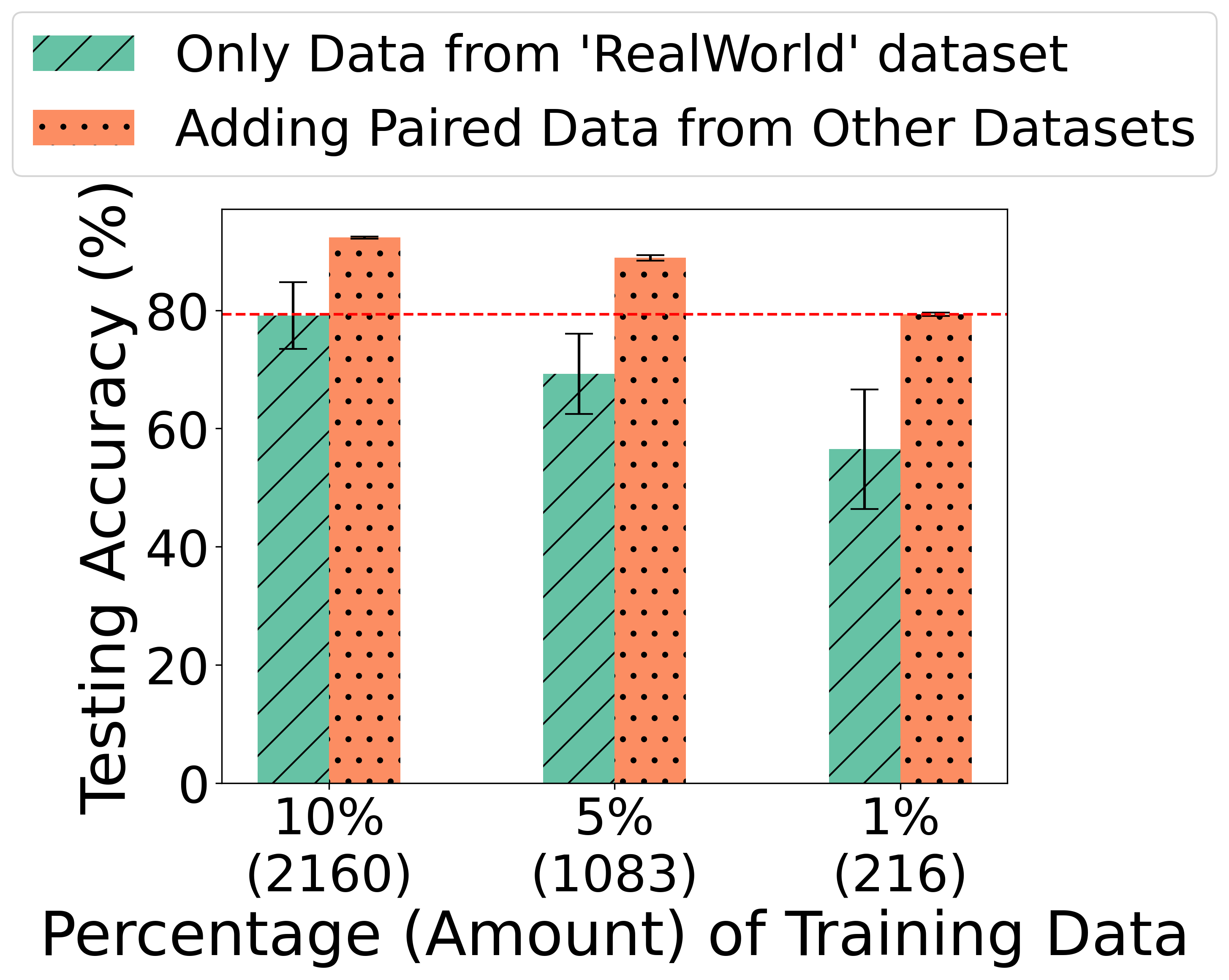}
     \caption{Acc binding.}
     \label{fig:cross-dataset-IMU-acc-bind}
    \end{subfigure}
    \caption{Performance of adding pseudo-paired data in cross-dataset binding for activity recognition.}%
    \vspace{-1em}
    \label{fig:cross-dataset-IMU}
\end{figure}

\begin{figure}
     \setlength{\abovecaptionskip}{0.cm}
    \setlength{\belowcaptionskip}{0.cm}
    \centering
  \begin{subfigure}{.49\linewidth}
    \setlength{\abovecaptionskip}{0.cm}
    \setlength{\belowcaptionskip}{-0.cm}
    \centering
     \includegraphics[width = \textwidth]{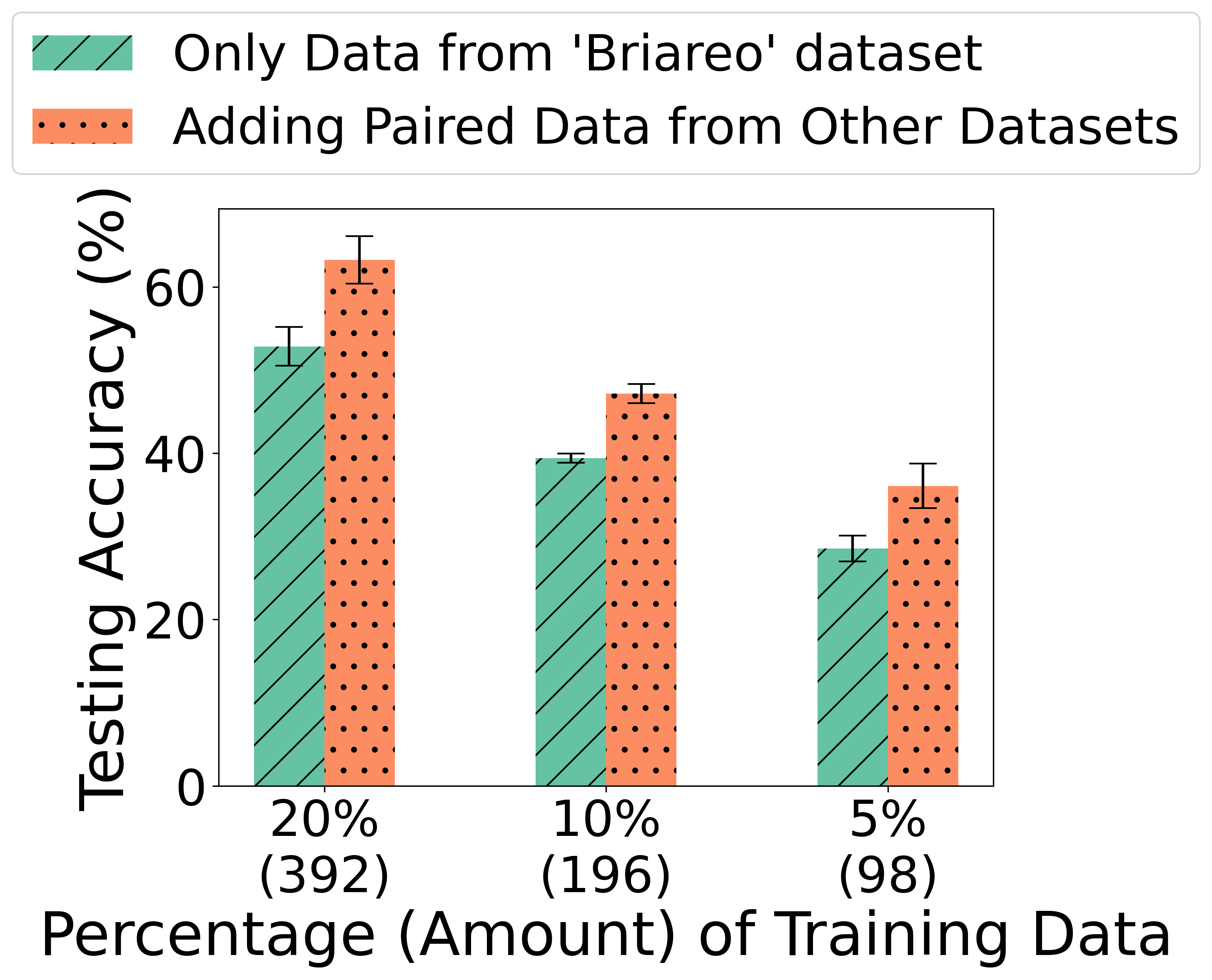}
    \caption{Label binding.}
      \label{fig:cross-dataset-gesture-label-bind}
  \end{subfigure}
      \begin{subfigure}{.49\linewidth}
    \setlength{\abovecaptionskip}{0.cm}
    \setlength{\belowcaptionskip}{-0.cm}
    \centering
     \includegraphics[width = \textwidth]{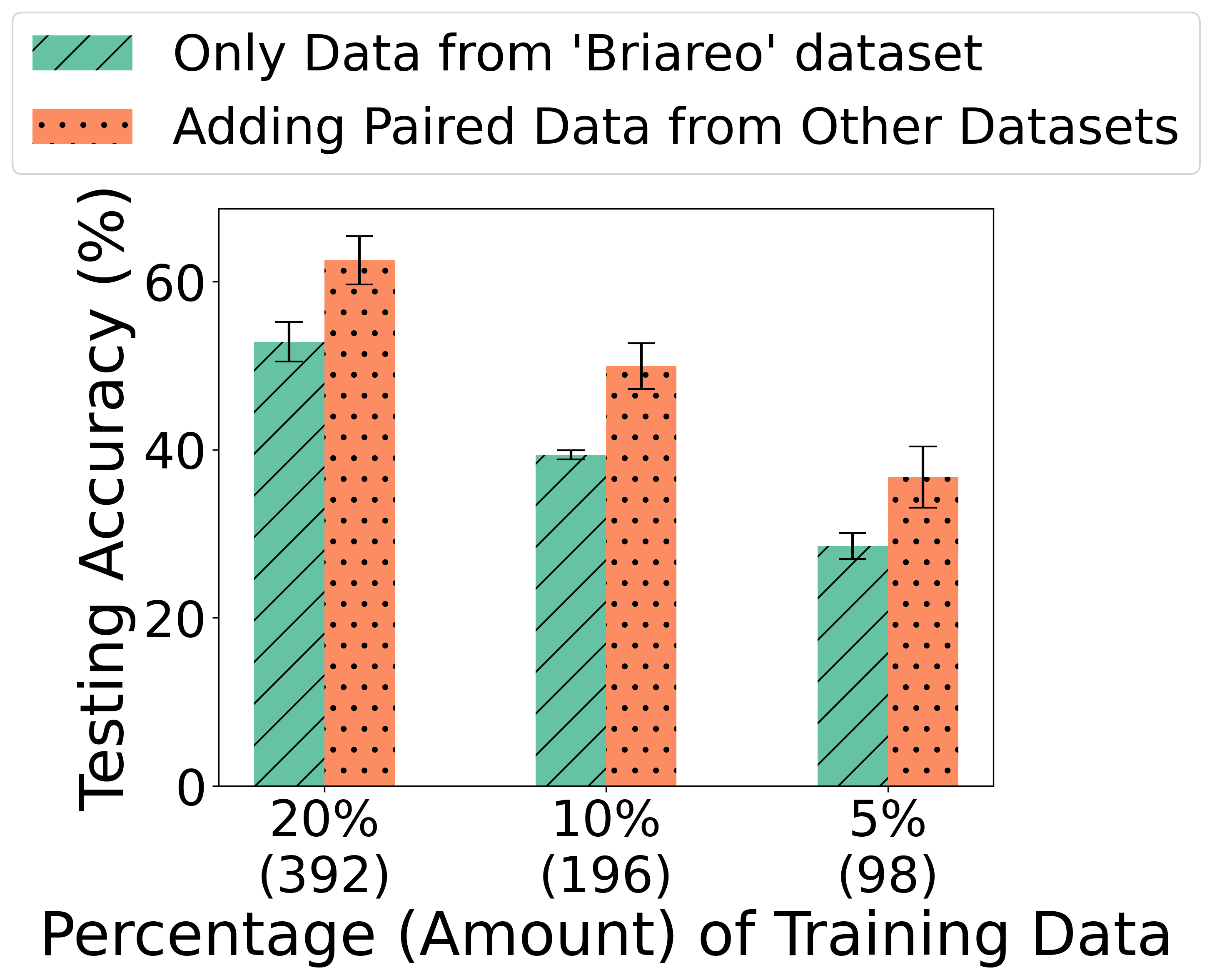}
     \caption{Skeleton binding.}
     \label{fig:cross-dataset-gesture-skeleton-bind}
    \end{subfigure}
    \caption{Performance of adding pseudo-paired data in cross-dataset binding for gesture recognition.}%
    \vspace{-1em}
    \label{fig:cross-dataset-gesture}
\end{figure}

\subsubsection{Performance in cross-dataset binding.} As discussed in Section \ref{sec:motivate_incomplete}, we can bind data from two different datasets to enhance the performance on a \emph{third distinct dataset} (after finetuning on the third). We evaluate this under two applications. For human activity recognition, we bind data (i.e., (Acc, Gyro, Mag)) from the MotionSense and Shoaib-right datasets, then finetune/test on RealWorld. We also test gesture recognition by binding data (i.e., (Skeleton, Depth, IR)) from GR4DHCI and DHG, then finetune/test on Briareo. Figure \ref{fig:cross-dataset-IMU} and \ref{fig:cross-dataset-gesture} showcase the performance on RealWorld and Briareo, respectively, with different proportions of finetuning data. We observe consistent improvements from adding pseudo-paired data across both label and sensor binding, particularly when finetuning with limited data. For example, on the RealWorld dataset, pre-training the model with pseudo label-paired data samples significantly boosts performance with only 1\% training data (77.9\%), which approaches the performance of using 10\% training data (79.1\%). Therefore, \name performs well when binding different datasets with various protocols, subjects, and environments.

\subsubsection{Performance with different levels of domain shifts.} In Table \ref{table:domain_shift}, we showcase \name's performance in the presence of varying domain shifts between the two training datasets that are bound together. We utilize the different environments naturally present in the MM-Fi dataset to emulate various levels of domain shift. In the case of ``No Domain Shift'', each binding dataset contains an equal number of unique subjects from Environment 1 and 2. With ``Medium Domain Shift'', the first dataset will have 80\% of its subjects from Environment 1, with the rest from Environment 2, with a similar allocation for the second dataset. Finally, with ``Large Domain Shift'', all subjects in the first dataset are from Environment 1, and all subjects in the second dataset are from Environment 2. We can see that \name retains its advantage over the baselines regardless of domain shift, exemplifying the utility of this approach.  


\begin{table}
\resizebox{\columnwidth}{!}
{%
\begin{tabular}{c|ccccccccc}
\thickhline
\makecell{Domain\\ Shift} & \makecell{Lower \\ Bound} & {Unimodal} & {MIM} & {MPM} & {CMG} & {DCM} & \makecell{Image\\Bind} & {\textbf{MMBind}} & \makecell{Upper\\ Bound} \\ 
\hline
None  & 68.43 & 33.55 & 72.94 & 64.46 & 73.21 & 71.72 & 73.76 & \textbf{79.48} & 78.68 \\ 
Medium & 67.69 & 55.58 & 72.30 & 72.46 & 72.66 & 73.81 & 72.61 & \textbf{78.48} & 75.14 \\ 
Large & 68.13 & 62.41 & 66.11 & 76.00 & 75.81 & 72.57 & 72.25 & \textbf{79.40} & 78.37 \\ 
\thickhline
\end{tabular}
}
\caption{Accuracy under varying domain shifts.}
\vspace{-2em}
\label{table:domain_shift}
\end{table}

}

\begin{table}
    \centering
 \setlength{\abovecaptionskip}{0.cm}
    \setlength{\belowcaptionskip}{0.cm}
    \resizebox{\linewidth}{!}{
     \begin{tabular}{c|c|c|c|c}
      \toprule
     Settings  & No Pairing & \makecell{MotionSense \\+ Shoaib-right}  & \makecell{MotionSense \\+ Shoaib-right \\ + Shoaib-left} & \makecell{MotionSense \\+ Shoaib-right \\ + Shoaib-wrist} \\
      \hline
    Samples & 0 & 17136 & 21636 & 21636\\
    Accuracy & 56.49 & 79.30 & \textbf{81.69} & \textbf{78.89} \\
      \bottomrule
    \end{tabular}}
    \caption{Impact of paring more than two datasets.}
    \vspace{-1em}
    \label{table:pairing_more_dataset}
\end{table}

\subsubsection{Impact of pairing more than two datasets.} \label{sec:exp_multiple_datasets}Table \ref{table:pairing_more_dataset} illustrates the performance of binding more than two datasets. While adding pseudo-paired data typically enhances performance, the improvement does not always correlate with the number of paired samples. For instance, {binding data from the left pocket and right pocket improves performance (MotionSense + Shoaib-right + Shoaib-left), while binding data from the wrist and right pocket results in a slight performance drop (MotionSense + Shoaib-right + Shoaib-wrist)}. This may be attributed to the significant differences in data distribution between the Shaib-wrist and Shaib-right datasets, which arise from variations in sensor placement. {Therefore, MMBind is effective when sensor locations are similar and do not affect the data distributions of the same events.} Moreover, when pairing multiple datasets, it is essential to consider the heterogeneity of the different paired datasets.



\subsection{Understanding MMBind's Performance}
We conduct an ablation study on the UTD dataset with Acc sensor binding to understand the effectiveness of \name. The results on other datasets are similar.

\subsubsection{Ablation study} Figure \ref{fig:ablation_study} compares the mean accuracy with different design components of \name. We evaluate the impact of three components: leveraging paired data (denoted as C1), adaptive multimodal learning with heterogeneous data (C2) and weighted contrastive learning based on pairing similarity (C3). The mean accuracy increases as more components are added while maintaining a significan t advantage over the lower bound (over 30\%), validating the effectiveness of different components in \name's design.

\subsubsection{Effectiveness of data pairing.} 
\begin{figure}
     \setlength{\abovecaptionskip}{0.cm}
    \setlength{\belowcaptionskip}{-0.cm}
    \centering
    \begin{subfigure}{.47\linewidth}
    \setlength{\abovecaptionskip}{0.cm}
    \setlength{\belowcaptionskip}{-0.cm}
    \centering
     \includegraphics[width = \linewidth]{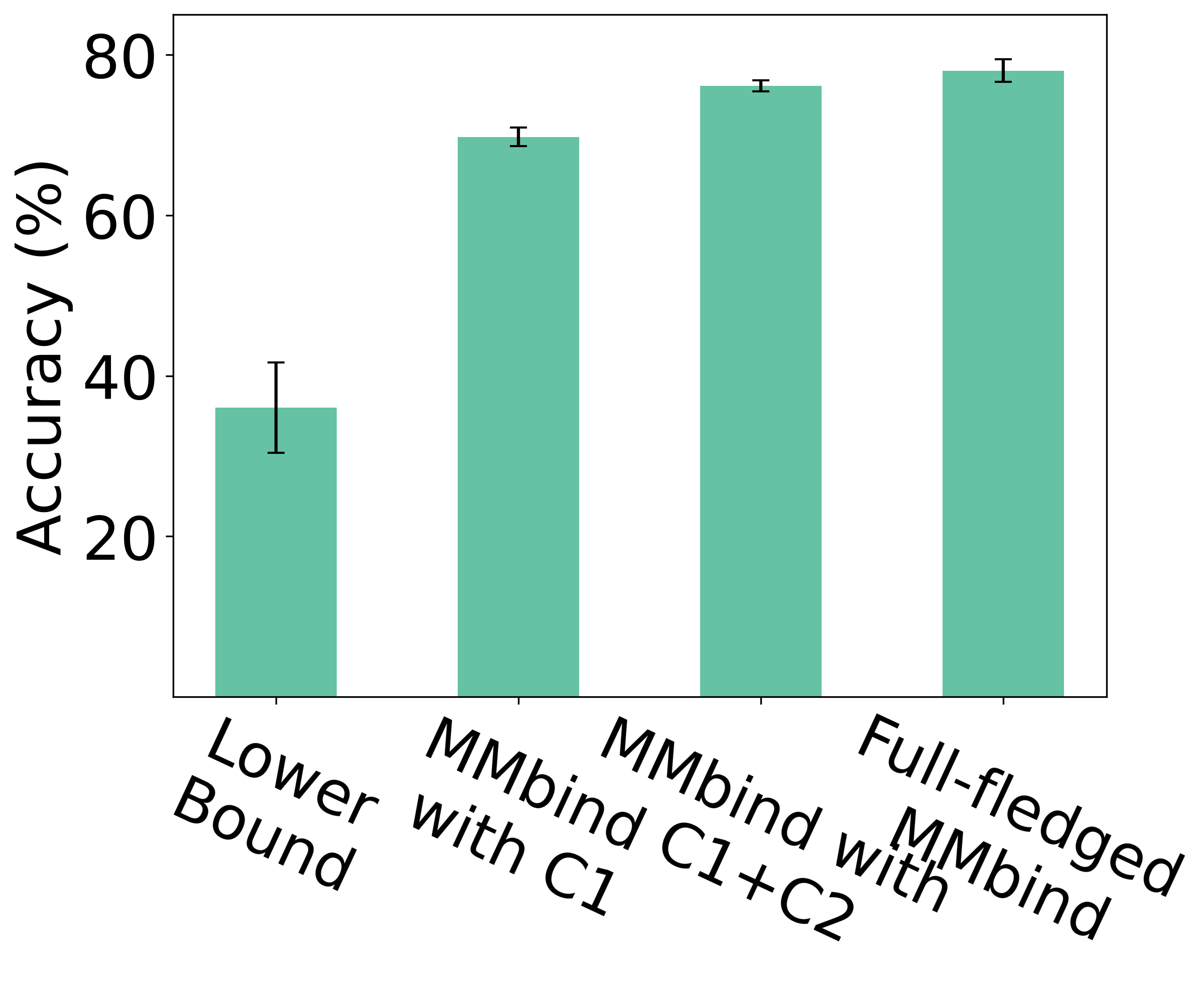}
     \caption{Ablation study.}
     \label{fig:ablation_study}
    \end{subfigure}\hspace{2pt}
  \begin{subfigure}{.48\linewidth}
    \setlength{\abovecaptionskip}{-0.cm}
    \setlength{\belowcaptionskip}{-0.cm}
    \centering
	\includegraphics[width=\linewidth]{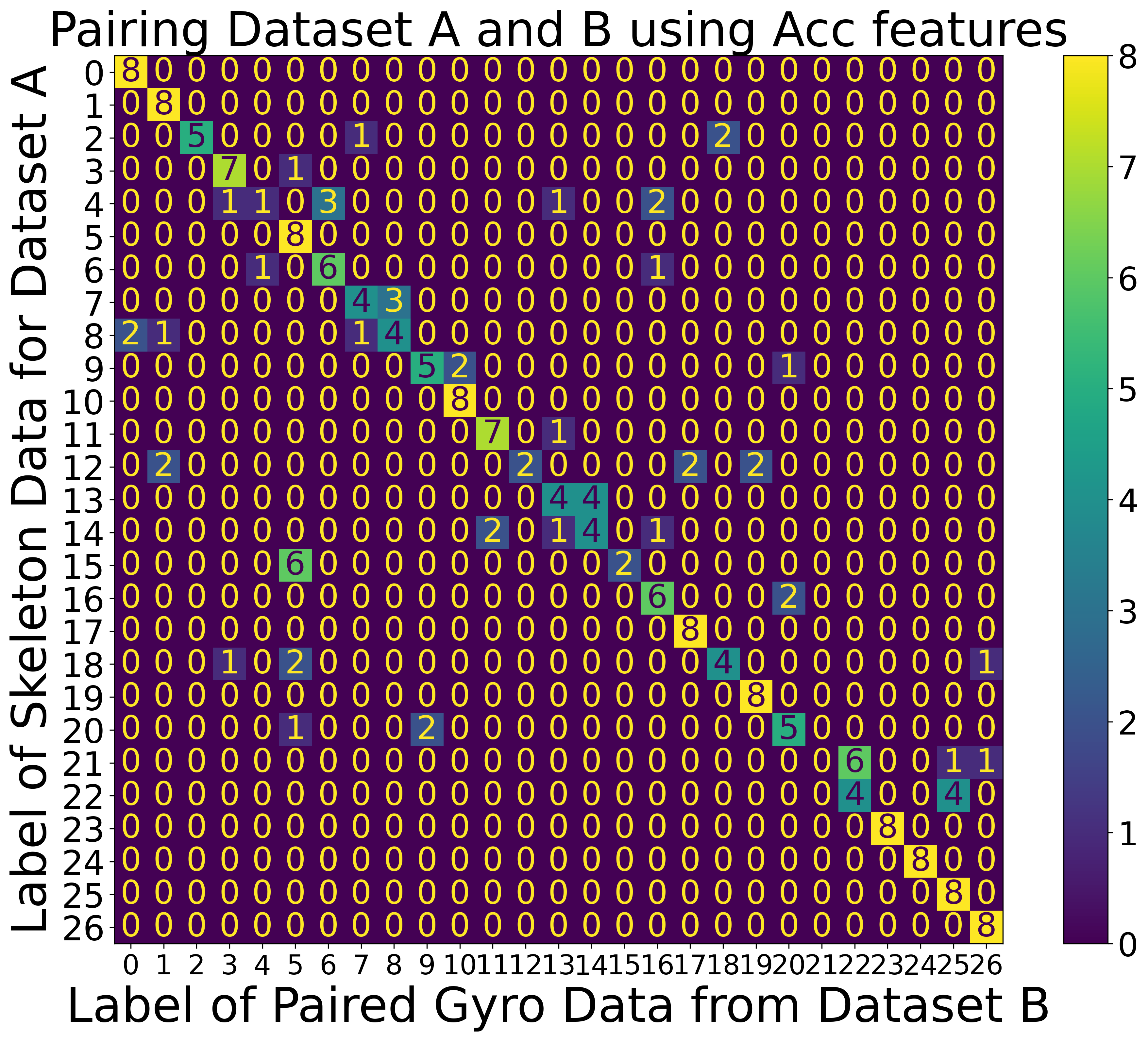}
	\caption{Data pairing.}
	\label{fig:data_pairing}
  \end{subfigure}
    \caption{Understanding \name's Performance.}
    \vspace{-1em}
\end{figure}

Figure~\ref{fig:data_pairing} visualizes the results of data pairing using Acc data as the shared modality. The x-axis and y-axis represent the indices of class labels for paired data samples from two incomplete datasets, respectively. The diagonal values indicate the number of correctly paired data samples that share the same label. Overall, \name achieves 69.76\% pairing accuracy, demonstrating the effectiveness of its data pairing design.

\subsection{Micro-benchmark Performance}

\subsubsection{Selecting the binding modality} 
\label{sec:exp_select_common_modality}
We study the impact of different binding modalities. Using the MM-Fi dataset, we compare the effectiveness of binding radar and depth data with either skeleton or WiFi CSI data. 
In contrast, using skeleton data as the binding modality significantly improves both pairing accuracy and multimodal performance.
Table \ref{table:diff_binding_modality} shows the data pairing accuracy (proportion of same-class data pairs) and the resulting performance. Pairing accuracy with WiFi is only 4.2\%, slightly better than random chance pairing. As a result, multimodal learning with WiFi-paired data achieves only 68.79\% accuracy, with minimal improvement over utilizing randomly paired samples.
These results confirm our analysis in Section \ref{sec:select_common_modality} that the common modality must effectively distinguish instances of each class.




\begin{table}
    \centering
    \begin{minipage}[t]{0.48\linewidth}
        \centering
    \resizebox{\linewidth}{!}{
     \begin{tabular}{c|cc}
          \thickhline
         \makecell{Binding\\Modality}  & \makecell{Pairing\\Accuracy}  & \makecell{\name\\ Performance} \\
          \midrule
         Skeleton &  \textbf{77.55\%} & \textbf{77.72\%} \\
        WiFi CSI & 4.2\% & 68.79\% \\
         Randomly & 2.31\% & 68.96\% \\
          \thickhline
        \end{tabular}
        }
        \caption{Different binding modalities.}
        \label{table:diff_binding_modality}
    \end{minipage}
    \hspace{0.5pt} 
    \begin{minipage}[t]{0.49\linewidth}
        \centering
        \resizebox{\linewidth}{!}{
        \begin{tabular}{c|cc}
              \thickhline
              \makecell{Amount of\\Paired Data} & \makecell{Pairing\\Accuracy} & \makecell{\name\\Performance} \\
              \midrule
             169 samples &  82\% & 72.04\% \\
            432 samples & 77\% & 76.32\% \\
            2,610 samples & 40\% & 75.89\% \\
              \thickhline
            \end{tabular}
        }
    \caption{Impact of paired dataset size.}
    \label{table:diff_envs}
    \end{minipage}
    \vspace{-2em}
\end{table}

\subsubsection{Different pairing schemes} \label{sec:evaluation_pairing_data}
During data pairing, we can either pair each sample with the one having the highest feature similarity or with multiple samples of lower similarity. We explore this trade-off between \emph{the size of the pseudo-paired dataset} (more paired data) and \emph{the pairing accuracy} (more accurately paired data) using the MM-Fi dataset. The 432 samples were obtained by pairing most similar samples, while the 169 and 2,610 samples were obtained by setting various thresholds on similarity.
Table \ref{table:diff_envs} shows that downstream accuracy will suffers at both extremes. Therefore, in \name, we choose the scheme of highest-similarity pairing.


\begin{figure}
     \setlength{\abovecaptionskip}{0.cm}
    \setlength{\belowcaptionskip}{-0cm}
    \centering
    \begin{minipage}{.52\linewidth}
    \setlength{\abovecaptionskip}{0.cm}
    \setlength{\belowcaptionskip}{-0.cm}
    \centering
	\includegraphics[width=0.95\linewidth]{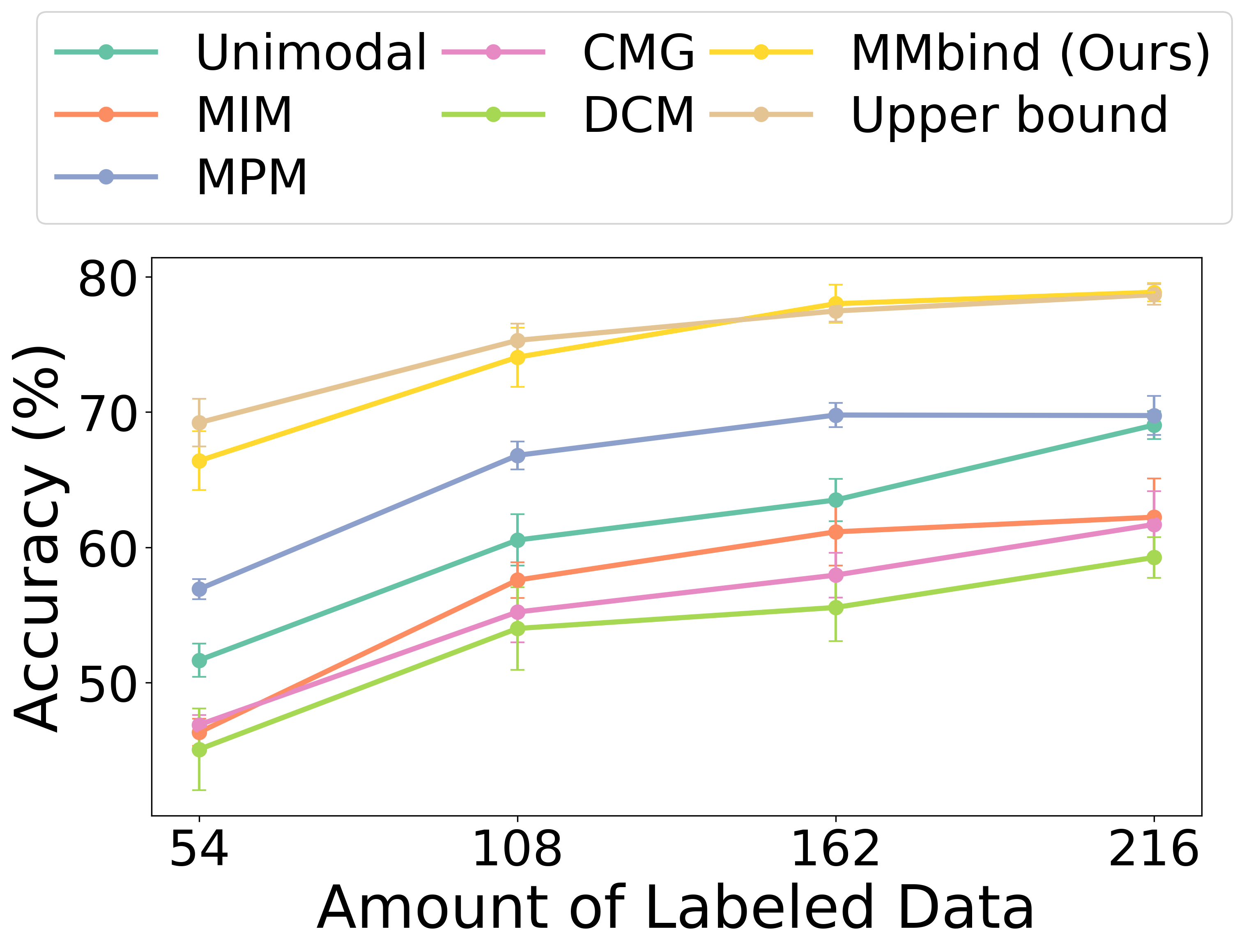}
	\caption{Different amounts of naturally paired data.}
	\label{fig:labels-all-acc-bind}
    \end{minipage}
    \hspace{0.5pt}
  \begin{minipage}{.45\linewidth}
    \setlength{\abovecaptionskip}{0.cm}
    \setlength{\belowcaptionskip}{-0.cm}
    \centering
	\includegraphics[width=\linewidth]{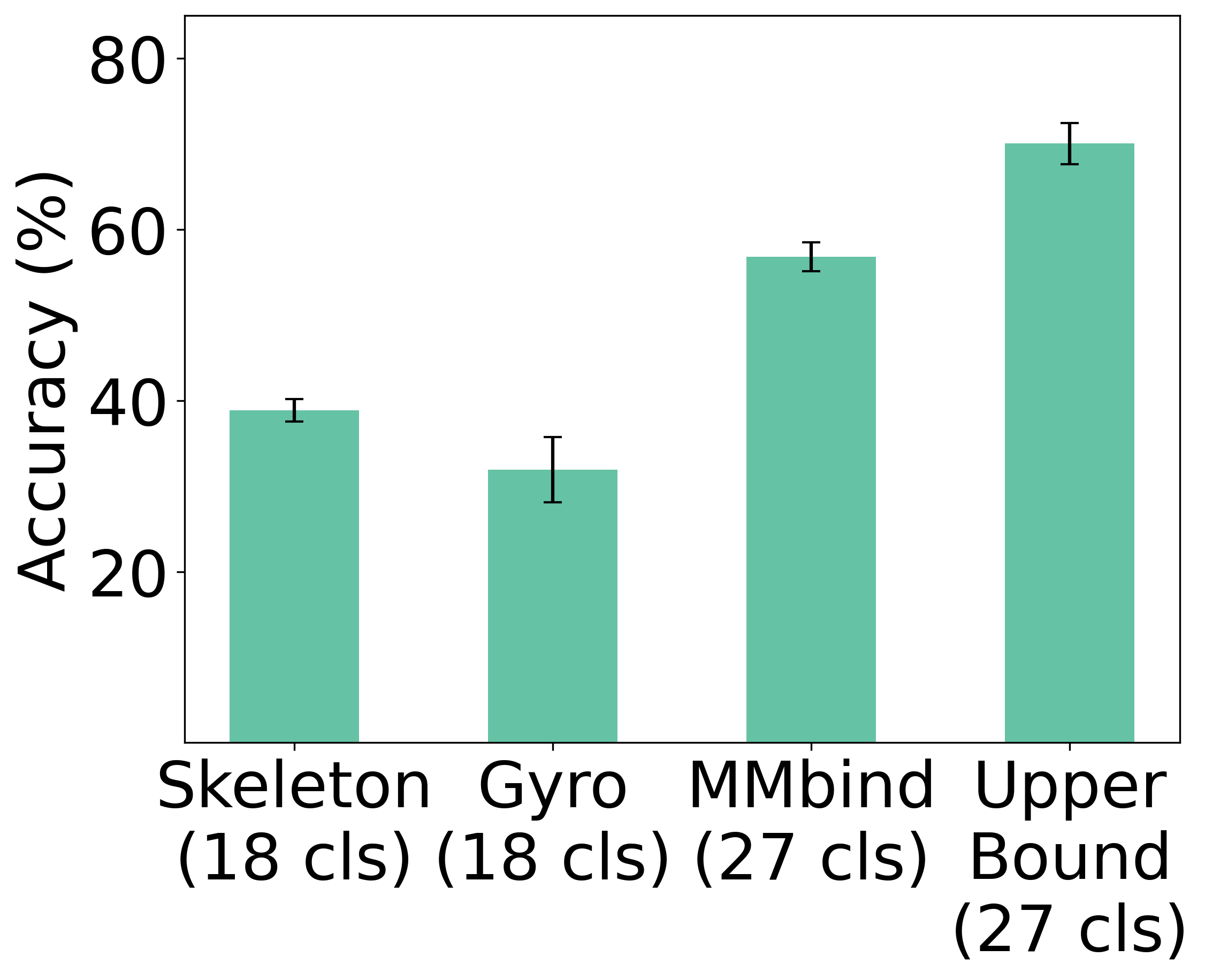}
	\caption{Partial class overlap.}
	\label{fig:UTD-partial-label-binding}
  \end{minipage}
  \vspace{-2em}
\end{figure}


\subsubsection{Different amounts of naturally paired data} We compare the accuracy performance using varying amounts of naturally paired data while maintaining the same set of incomplete training data samples. As shown in Figure~\ref{fig:labels-all-acc-bind}, all approaches achieve higher accuracy with more naturally paired data. Moreover, \name consistently outperforms other baselines across various settings, which shows the superior generalization ability of the multimodal embeddings learned by \name.



\subsubsection{Partial class overlap between different datasets.} We evaluate the performance of \name on label binding when there is partial class overlap between different datasets. Specifically, using the UTD dataset, we configure the unimodal skeleton data and Gyro data to have 18 classes, with only 10 classes overlapping. As shown in Figure~\ref{fig:UTD-partial-label-binding}, unimodal learning exhibits poor performance due to covering only partial classes. However, by pairing data from similar classes, \name can generate pseudo-paired data from 27 classes, resulting in a significant performance improvement.




\subsection{System Evaluations}

\subsubsection{Training overhead on edge} 
\begin{figure}
     \setlength{\abovecaptionskip}{0.cm}
    \setlength{\belowcaptionskip}{-0cm}
    \centering
    \begin{minipage}{.45\linewidth}
    \setlength{\abovecaptionskip}{0.cm}
    \setlength{\belowcaptionskip}{-0.cm}
    \centering
	\includegraphics[width=\linewidth]{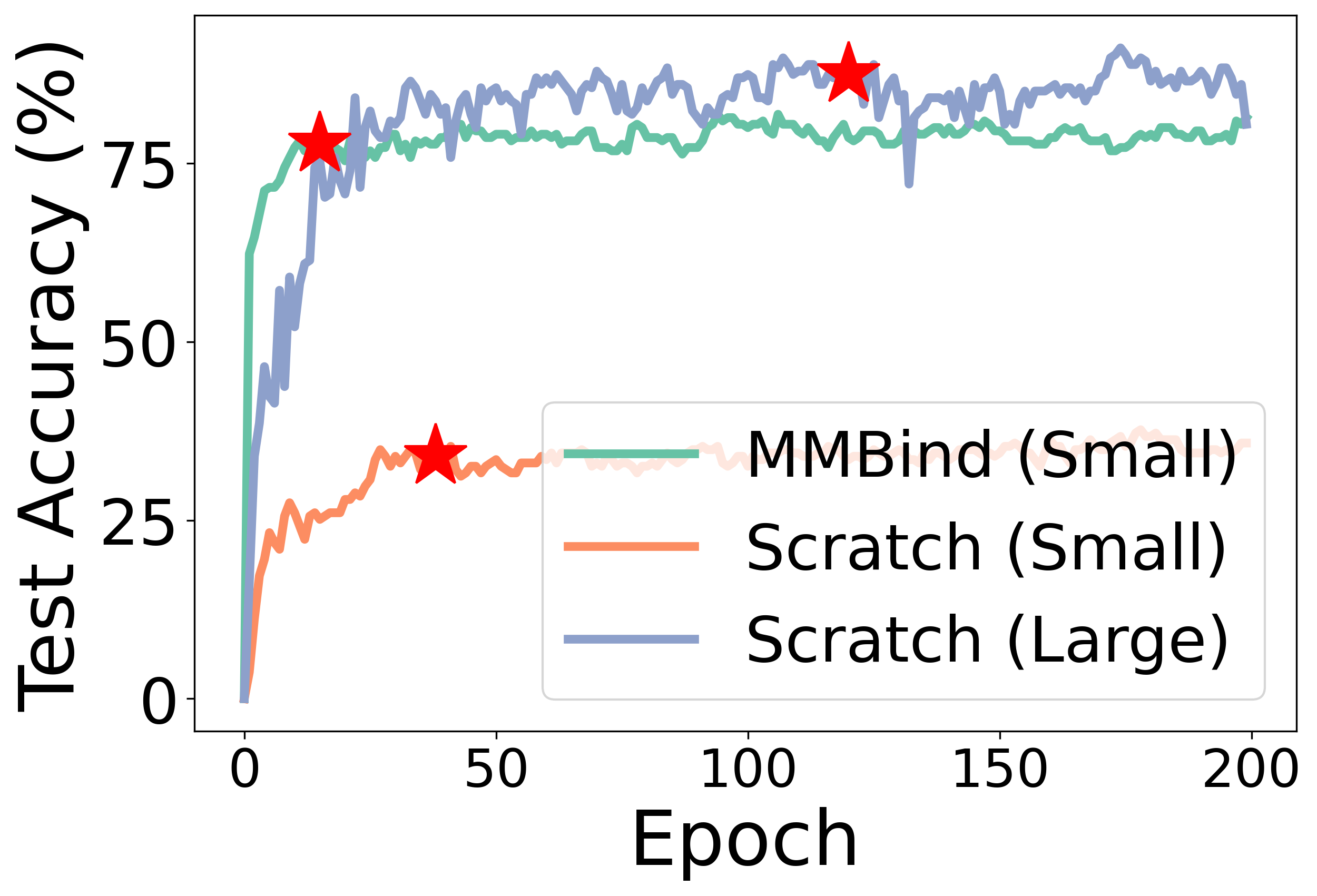}
	\caption{Convergence of edge training.}
	\label{fig:Tx2_convergence}
    \end{minipage}
    \hspace{1pt}
  \begin{minipage}{.52\linewidth}
    \setlength{\abovecaptionskip}{0. cm}
    \setlength{\belowcaptionskip}{-0. cm}
    \centering
	\includegraphics[width=\linewidth]{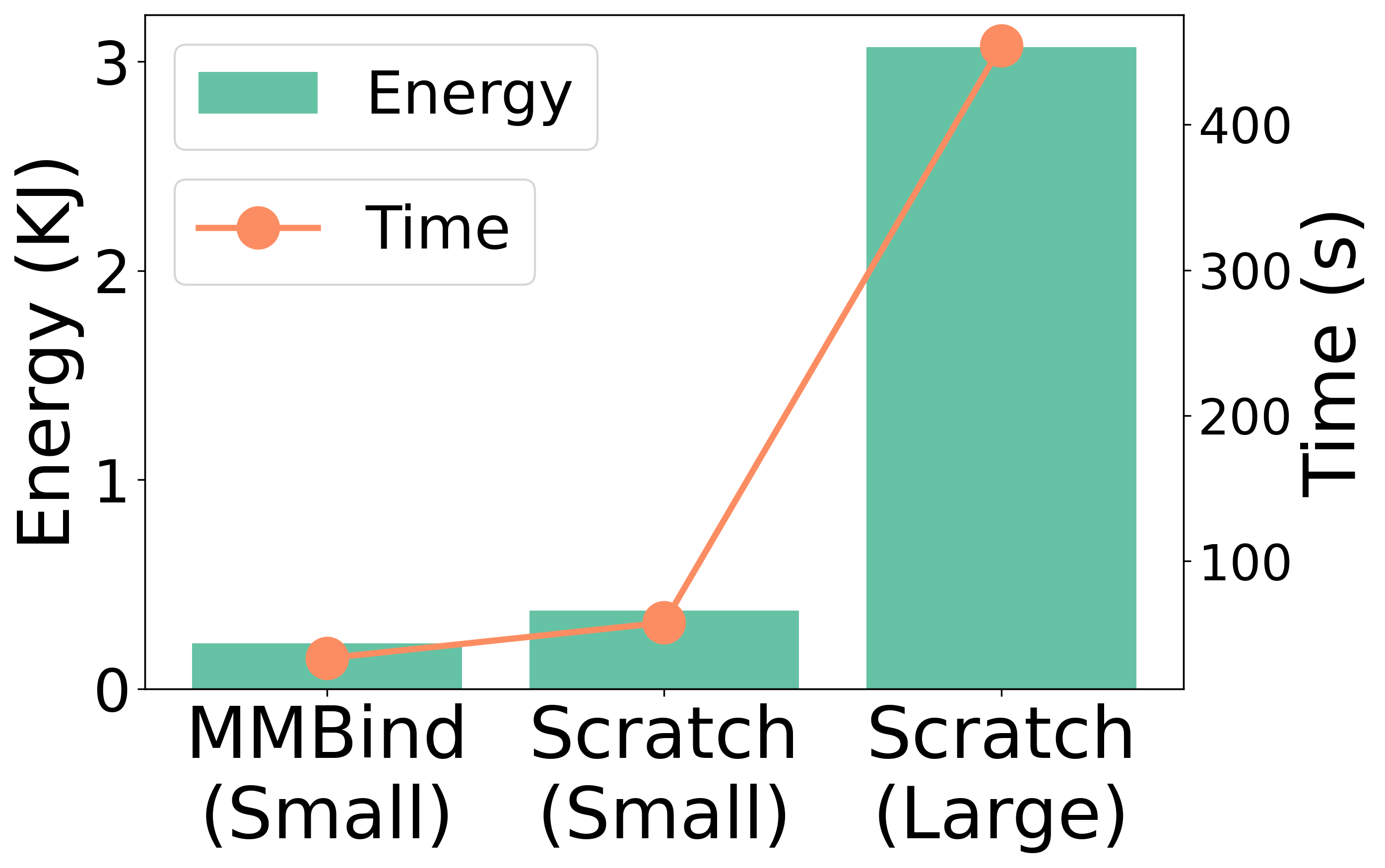}
	\caption{Time and power consumption on Nvidia Tx2.}
	\label{fig:Tx2-power}
  \end{minipage}
  \vspace{-1em}
\end{figure}

We now evaluate the training overhead of \name. We first perform pre-training with incomplete data on the server and then fine-tune the model using small paired data on NVIDIA Jetson TX2. We compare the training overhead to training from scratch using small and large paired datasets, respectively.
As shown in Figure \ref{fig:Tx2_convergence}, \name converges faster than the baselines while enhancing performance with limited paired data, thanks to its ability to leverage incomplete data for pre-training. Additionally, Figure \ref{fig:Tx2-power} further shows that \name incurs lower time and power consumption during the fine-tuning stage, making it feasible for resource-limited edge devices.


\begin{table}
    \centering
    \resizebox{\linewidth}{!}{
     \begin{tabular}{c|ccccccc}
          \thickhline
         Modality  & A  & G & S & A+G & A+S & \textbf{S+G} & A+S+G\\
          \midrule
         Lower bound &  60.04\% & 63.35\% & 	30.82\% & 	63.80\% & 	39.54\% & 	40.05\% & 49.00\% \\
        MMBind & 68.70\% & 69.96\% & 	66.42\% & 	74.78\% & 	78.54\% & 	79.61\% & 	77.90\% \\
         Upper Bound & 72.20\% &	73.39\% & 82.58\% &	76.44\% &	84.25\% &	86.45\% &	86.21\% \\
          \thickhline
        \end{tabular}
        }
        \caption{Robustness to different binding modalities in inference. Pre-trained on paired (Skeleton, Gyro).}
        \vspace{-2em}
        \label{table:inference_diff_modality}
\end{table}

\subsubsection{Deployment robustness.} \label{sec:deployment_robustness} We now evaluate the robustness of the multimodal embedding trained by \name. 
Table \ref{table:inference_diff_modality} shows the performance of \name when deployed on various modality combinations during inference. The model is pre-trained with paired data from (Skeleton, Gyro) and original (Acc, Gyro) and (Acc, Skeleton) data, and fine-tuned on limited labeled data with different modality combinations. In addition to the data combinations used during pre-training, \name significantly improves over the lower bound on unseen modality combinations, such as unimodal data (Acc, Gyro, Skeleton) and fully-modality (Acc + Gyro + Skeleton). This indicates that the multimodal embeddings learned by \name are robust across different modality combinations.


\section{Discussion}

\textbf{Advanced definition of shared modalities.} Beyond sensor data and labels, we can further refine how \name defines the shared modalities to utilize various types of common information across different datasets. For example, the binding modalities could include data from identical sensor types deployed at different locations, different sensor types (e.g., RGB and Depth), and even timestamp and location information. 

\textbf{Beyond pairwise matching.} In \name, we construct pseudo-paired multimodal samples by matching the most similar samples between disparate datasets. However, as shown in Section \ref{sec:evaluation_pairing_data}, there is a trade-off between more accurate pairing (pairing the most similar sample) and a larger, but potentially noisier, set of pairings (pairing multiple similar samples). In the future, we can explore advanced pairing schemes, such as using a weighted sum of multiple similar samples to form pseudo-paired data.

\textbf{Augmenting label binding.} Although we can leverage LLMs to understand data labels for pairing, text labels as a sensing modality often depend on the specific downstream task. For example, if we pair disparate sensor data using activity labels but deploy the pre-trained multimodal model for user identification, performance may degrade. In the future, we will explore augmenting data labels with diverse contextual information (e.g., subjects, devices, environments) for pairing, enhancing the model's ability to generalize across various downstream tasks.

\textbf{Advancing multimodal foundation model for IoT.} By applying \name for cross-dataset binding, we demonstrate its potential for generating foundational datasets for IoT by effectively leveraging large amounts of incomplete datasets that contain subsets of the required modalities. This approach creates valuable opportunities for training multimodal foundation models in the IoT domain. 


\section{Conclusion}
In this paper, we introduce \name, a new framework for multimodal learning on \emph{distributed and heterogeneous IoT data} through binding them with the shared modalities.
By incorporating diverse modalities like sensor data or labels to bind disparate and incomplete data, \name shows potential for advancing multimodal foundational model training in IoT applications. 
We also provide insights into the optimal conditions for adopting \name and discuss future directions to further enhance the performance of \name.

\begin{acks}
The research reported in this paper was sponsored in part by the DEVCOM Army Research Laboratory under award \#W911NF-17-2-0196, the National Institutes of Health under award \#1P41EB028242, and the Air Force Office of Scientific Research under awards \# FA95502210193 and FA95502310559. Any findings in this material are those of the author(s) and do not reflect the views of any of the above funding agencies. The U.S. Government is authorized to reproduce and distribute reprints for Government purposes notwithstanding any copyright notation here on.
\end{acks}

\bibliographystyle{ACM-Reference-Format}
\bibliography{reference}

\end{document}